\documentclass[NewProceedings,letterpaper]{ascelike-new}
\usepackage{float}
\usepackage{comment}
\usepackage[utf8]{inputenc}
\usepackage[T1]{fontenc}
\usepackage{lmodern}
\usepackage{graphicx}
\usepackage{ragged2e}
\usepackage[figurename=Fig.,labelfont=bf,labelsep=period]{caption}
\usepackage{amsmath}
\usepackage{xcolor} 
\usepackage{newtxtext,newtxmath}
\usepackage{caption}
\usepackage[colorlinks=true,citecolor =red,linkcolor=black]{hyperref}
\usepackage[hyphenbreaks]{breakurl}
\usepackage[export]{adjustbox}
\usepackage{fancyhdr}
\usepackage{comment}

\usepackage{tabularx}
\usepackage{longtable}
\usepackage{booktabs} 

\renewcommand{\refname}{References} 

\usepackage{subcaption}
\usepackage[capitalise]{cleveref}

\usepackage{natbib} 
\let\cite\citep 

\usepackage{etoolbox}
\apptocmd{\thebibliography}{\setlength{\itemsep}{0pt}}{}{}

\usepackage{titlesec}
\titlespacing*{\section}
{0pt}{3.5ex plus 1ex minus .2ex}{2.3ex plus .2ex}
\titlespacing*{\subsection}
{0pt}{3.25ex plus 1ex minus .2ex}{1.5ex plus .2ex}
\titlespacing*{\subsubsection}
{0pt}{3.25ex plus 1ex minus .2ex}{1.5ex plus .2ex}

\titleformat{\section}
  {\normalfont\Large\bfseries}{\thesection}{1em}{}
\begin{document}

\title{Physics-Inspired Deep Learning and Transferable Models for Bridge Scour Prediction}

\author[1]{Negin Yousefpour, PhD, PE}
\author[2]{Bo Wang, PhD}

\affil[1]{Department of Infrastructure Engineering, The University of Melbourne. Email: negin.yousefpour@unimelb.edu.au}
\affil[2]{Department of Infrastructure Engineering, The University of Melbourne. Email: bo.wang@unimelb.edu.au}

\maketitle

\begin{abstract} {This paper introduces scour physics-inspired neural networks (SPINNs), a hybrid physics-data-driven framework for bridge scour prediction using deep learning. SPINNs integrate physics-based, empirical equations into deep neural networks and are trained using site-specific historical scour monitoring data. Long-short Term Memory Network (LSTM) and Convolutional Neural Network (CNN) are considered as the base deep learning (DL) models. We also explore transferable/general models, trained by aggregating datasets from a cluster of bridges, versus the site/bridge-specific models. Despite variation in performance, SPINNs outperformed pure data-driven models in the majority of cases. In some bridge cases, SPINN reduced forecasting errors by up to 50 percent. The pure data-driven models showed better transferability compared to hybrid models. The transferable DL models particularly proved effective for bridges with limited data. In addition, the calibrated time-dependent empirical equations derived from SPINNs showed great potential for maximum scour depth estimation, providing more accurate predictions compared to commonly used HEC-18 model. Comparing SPINNs with traditional empirical models indicates substantial improvements in scour prediction accuracy. This study can pave the way for further exploration of physics-inspired machine learning methods for scour prediction.}
\end{abstract}

\setcounter{section}{0}
\section{Introduction} \label{sec-introduction}
\noindent
Scour is recognized as the leading cause of bridge failure in many countries. In the United States, scour is responsible for approximately 60\% of bridge failures \cite{BriaudJL2012UnknownScour}. Historical data from other countries and regions, such as Australia, Taiwan, Japan, Germany, France, and Iran, shows that a large number of bridges have suffered scour-related failures due to typhoons and floods in the past few decades \cite{Pizarro2020TheReview}.

The complexity of predicting the expected maximum depth of scour at bridge piers stems from the intricate interactions between soil, hydrodynamic forces and the pier structure. Key challenges include uncertainties in riverbed material, flow conditions, geomorphological variations, and the increasing impacts of climate change. In recent decades, numerous research efforts have focused on developing empirical models to estimate scour depth using laboratory experiments and field observations. Despite these efforts, many empirical models tend to overestimate or underestimate scour depth due to their limited generalization across diverse riverbeds, flow, and structural conditions \cite{sheppard2011scour}. For instance, the widely used HEC-18 scour equation, despite its evolution over the past two decades, has notable limitations, including insensitivity to various geological conditions, particularly to cohesive soils. Also, the empirical models are unable to provide a reliable assessment of scour variation with time and risk of scour failure, especially under flood conditions \cite{Hec-18:2012}. In response to this pressing need for more reliable predictive methods to enhance scour risk assessment and bridge safety, various Artificial Intelligence (AI) and Machine Learning (ML) methods have been explored. Readers are referred to \citet{yousefpour2021machine}, \citet{yousefpour2023towards}, \citet{hashem2024application}, and the cited references for a full literature review. 

The ML models have been shown to outperform traditional empirical equations in estimating maximum scour depth. However, their performance remains restricted within the domain of training data and is highly dependent on the data quality and size, which are often scarce and do not cover a wide range of geological, hydraulic, and geomorphological conditions. In addition, these models are typically designed to predict the maximum scour depth for a given flow discharge and are not suitable for dynamic real-time forecasting. Given the limitations of current scour prediction models, bridge authorities have turned to monitoring solutions, including the use of continuous remote sensing systems enabling more reliable risk management for critical and large-scale bridges \cite{NCHRP:2009,Briaud:2011}. Pioneering works by \citet{yousefpour2021machine}, \citet{yousefpour2023towards}, and \citet{hashem2024application} introduced deep learning (DL) solutions, specifically long-short-term memory (LSTM) networks and the convolutional neural network (CNN), for real-time scour forecasting. These models leverage LSTM and temporal CNN strength in temporal pattern recognition and capture the complex physical process of scour without direct feature extraction. By training on historical scour monitoring data, including time series of bed elevation, flow depth, and velocity, the DL models have shown the ability to predict upcoming scour depths up to a week in advance for case-study bridges in Alaska and Oregon.

Physics-based machine learning has accelerated as an emerging field in the past few years. The core idea is to incorporate physics law and constraints into the process of machine learning from data \cite{Karpatne2017,Karniadakis2021}. The hybrid physics-data-driven models have been introduced by a number of pioneering studies, referred to as physics-guided neural networks (PGNN) \cite{Jia2021Physics-GuidedProfiles}, physics-constrained neural networks \cite{Raissi2019}, and physics-informed neural networks (PINN) \cite{Zhu2019}. These terms have been used alternately in various recent studies \cite{Faroughi2024}. The incorporation of physical equations governing the input-response relationship, especially in the form of PDE/ODE within the loss function, is being increasingly referred to as physics-informed ML. PINNs have been particularly developed to solve PDEs in application to various problems in solid mechanics \cite{Haghighat2021}, porous media simulations \cite{Chen2023,Amini2023}, constitutive modelling, and soil consolidation \cite{Eghbalian2023,KaiQiLi2024,Lan2024,Tian2023,Masi2024}. Likewise PGNNs have found successful applications in structural analysis \cite{Yu2020,Chen2021}, geohazard assessment \cite{Pei2023,Zhang2020} tunneling \cite{YongshengLi2024}, fluid mechanics \cite{Yousif2022}, among other fields.

This paper introduces hybrid physics-data-driven models for scour prediction. A novel physics-inspired deep learning framework is developed by integrating semi-empirical scour models with prominent deep learning algorithms. As closed-form equations are used in this study, and to avoid misinterpretations, the term "physics-inspired neural networks" is adopted. We introduce SPINNs, \textit{Scour Physics-Inspired Neural Networks}, which leverage the temporal feature extraction capability of recurrent neural networks and convolutional neural networks, while respecting the physics driving the scour process. 

The SPINNs are trained using historical scour monitoring data (more than 15 years) from a number of case-study bridges in Alaska. As a by-product of SPINNs, new empirical equations are introduced that are calibrated through deep learning algorithms. These DL-calibrated empirical equations can have superior performance over locally calibrated empirical models (e.g., HEC-18). 

We also explore the viability of \textit{transferable} scour models; these general models are trained with collective monitoring data from a cluster of bridges within a region and implemented to predict scour for individual bridges. The transferability of the proposed data-driven and hybrid models are investigated across the case studies and their accuracy is compared to bridge/site-specific models. 

\section{Approach}\label{sec-approach}
\noindent

\subsection{Notations and Definitions}
\noindent
To provide better clarity to our methodology, notations are provided in Table~\ref{tab:table_notation}. 

\begin{table}[!ht]
    \centering
\caption{Notations.}
\label{tab:table_notation}
    \begin{tabular}{l l l}
    \toprule
        Symbol & Description & Unit  \\  
    \midrule
        $t$ & Timestep                                                                     & $hour$  \\
        $E_{stage}$ & Rive flow elevation measured by stage sensors at the upstream of the pier       & $m$     \\
        $E_{bed}$ &River bed elevation at the pier measured by Sonars                                & $m$     \\
        $E_{ref}$ & Reference elevation for calculating local scour depth                  & $m$     \\
        $y_{s}$ & Equilibrium scour depth                                                  & $m$     \\
        $y_{st}$ & Time-dependent scour depth                                              & $m$     \\
        $y_{1}$ & Flow depth directly upstream of the pier                                 & $m$     \\
        $M_{in}$  & Input length (timesteps) of the NN                                     & $N/A$     \\
        $M_{out}$ & Output length (timesteps) of the NN                                    & $N/A$     \\
        $Y_{s}$   & Sequential forecasting targets contains observed scour depth values    & $N/A$    \\
        $\hat{Y}$   & The prediction on $Y_{s}$                                            & $N/A$    \\
        $\hat{Y}_{s_{NN}}$ & The prediction on $Y_{s}$  by NN model                        & $N/A$     \\
        $\hat{Y}_{s_{PHY}}$ & The prediction on $Y_{s}$  by calibrated empirical equation  & $N/A$     \\
        $a$ & Pier width                                                                   & $m$     \\
        $L$ & Channel width                                                                & $m$     \\
        $V$ & Mean velocity of flow directly upstream of the pier                          & $m/s$   \\
        $g$ & Acceleration of gravity                                                      & $9.81m/s^2$ \\
        $A$ & Cross-sectional area of flow                                                       & $m^2$    \\
        $q$ & Average upstream discharge                                                   & $m^3/s$ \\
        $K_{1}$ & Correction factor for pier nose shape          &  $N/A$     \\
        $K_{2}$ & Correction factor for angle of attack of flow  &  $N/A$     \\
        $K_{3}$ & Correction factor for bed condition            &  $N/A$     \\
    \bottomrule
    \end{tabular}
\end{table}

\subsection{Scour Data and Preprocessing}
\noindent
The scour monitoring data of four bridges in Alaska is used to train the scour prediction models in this study. 
This data includes time series of the river bed, measured by sonar sensors at the bridge pier, and flow elevations, measured by stage sensors or gauges at the upstream, as well as river discharge, estimated based on measurement of flow velocity using velocimeter or Acoustic Doppler sensors. This data was provided by the Alaska Department of Transportation (DOT) and the US Geological Survey (USGS).

Some of the missing data, for example, the discharge data was obtained from the Alaska Science Center online platform\footnote{https://www.usgs.gov/centers/alaska-science-center/science/streambed-scour-bridges-alaska}. Most of the bridges on this platform contain the historical and real-time monitoring data collected from sonar and stage sensors attached to bridge piers with critical scour depth. The river discharge data in our methodology is only available for a few bridges at this point. Considering the data availability, we selected Bridge 212, 527, 539, and 742 with the key attributes presented in Table~\ref{tab:table_static_data}. The as-built bed elevation provides an important reference level to measure the scour depth. Pier and channel dimensions are listed in meters. Also, coefficients $K_1$, $K_2$, and $K_3$ are factors used in the HEC-18 empirical equations as defined in Table~\ref{tab:table_notation}.

This paper mainly follows the same preprocessing method outlined in \citet{yousefpour2021machine} and \citet{yousefpour2023towards} for the historical bridge scour monitoring dataset. The raw data is aggregated on an hourly basis. The key preprocessing steps include outlier removal, smoothing and denoising, and linear imputation of missing data. The data availability of each monitored attribute is impacted by many factors, such as sensor reliability, seasonal river freezing, and flood occurrences. We synchronised $E_{bed}$, $E_{stage}$, and $q$, ensuring the timesteps used for training the models include these three input features.

\begin{table}[ht]
\centering
\caption{Attributes of Case-Study Bridges.}
\label{tab:table_static_data}
\begin{tabular}{l l p{1.5cm} p{1cm} p{1cm} p{1cm} p{1cm} p{1cm} l l l}
    \toprule
    No &
    Name &
    As-built Bed-Elevation (m) &
    Channel Width (m) &
    Pier Width (m) &
    Pier Length (m) &
    Attack Angle (°) &
    Pier Nose Shape &
    $K_1$ &
    $K_2$ &
    $K_3$ \\
    \midrule
    212 & Kashwitna   River & 48.8   & 65.2  & 1.5  & 8.5  & 0  & Sharp & 0.9 & 1       & 1.1 \\
    527 & Salcha   River    & 193.2 & 153.3  & 1.2  & 10.7 & 10 & Round & 1   & 1.8     & 1.1 \\
    539 & Knik   River      & 10.4  & 152.7  & 1.3  & 7.9  & 0  & Sharp & 0.9 & 1       & 1.1 \\
    742 & Chilkat   River   & 35.2  & 152.4  & 1.5  & 7.3  & 0  & Round & 1   & 1       & 1.1 \\
    \bottomrule
    \end{tabular}
\end{table}

\subsection{Physics-Inspired Deep Learning Framework}
\noindent

\begin{figure}[!h]
    \centering
    \includegraphics[width=0.7\linewidth]{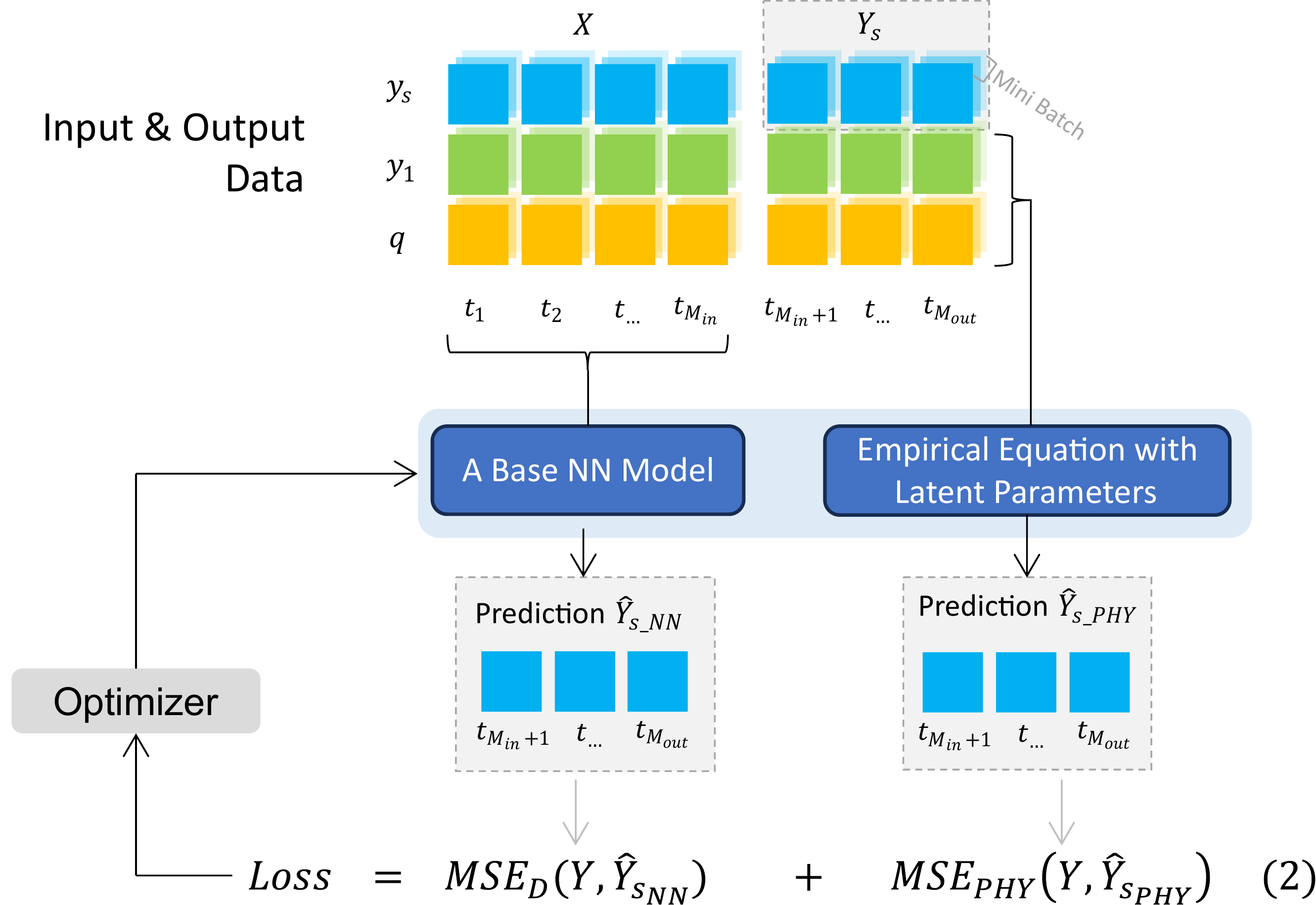}
    \caption{SPINN framework, illustrating the integration of physics-based empirical equations as additional loss terms in the neural network training process.}
    \label{fig:1}
\end{figure}

We introduce Scour Physics-inspired Neural Network (SPINN) as a new framework for scour prediction. This framework integrates an additional physical loss term using physics-based empirical equations. Over the past few decades, various semi-empirical equations have been developed to estimate local scour depth based on physical factors such as flow depth, flow velocity, pier geometry, and riverbed material/geology, calibrated based on laboratory and field testing \cite{sheppard2011scour,Hec-18:2012,kirby2015manual}. The overall architecture of SPINN is depicted in Fig.~\ref{fig:1}.


The SPINN models are developed based on our pure data-driven LSTM and CNN scour forecast algorithms developed in \citet{yousefpour2023towards} and \citet{hashem2024application} for scour forecast based on historical sonar, stage, and discharge time series data.
The SPINN model receives the time series data in a number of small slices each including an input part and output part. The input sequence $X$, includes scour depth, $y_s$, flow depth, $y_1$, and discharge, $q$ over $M_{in}$ timesteps, while the output sequence, $Y_s$ contains only scour depth, $y_s$ over $M_{out}$ timesteps. The discrepancy (loss) between the target (actual) scour depth ($y_i$) and the predicted scour depth ($\hat{y}_i$) is quantified over the output sequence, using the Mean Squared Error (MSE). As opposed to pure data-driven ML, the objective function is defined as the sum of two loss terms, a data-driven loss ($MSE_D$) and a physics-based loss ($MSE_{PHY}$), as described in Equations \ref{eq:mse} and \ref{eq:1}: 

\begin{equation} \label{eq:mse}
    MSE = \sum_{i=0}^{M_{out}}(y_i-\hat{y}_i)^2
\end{equation}

\begin{equation} \label{eq:1}
    Loss = MSE_D(Y,\hat{Y}_{s_{NN}}) + MSE_{PHY}(Y,\hat{Y}_{s_{PHY}})
\end{equation}

Within this framework, $\hat{Y}_{s_{NN}}$ represents the output from the base NN model (pure ML), and $\hat{Y}_{s_{PHY}}$ denotes the predictions yielded by the empirical equation (calibrated through ML training). For this purpose, the $y_1$ and $q$ within the output timesteps ($M_{out}$) are used as input data for the physics empirical equation. By concurrently optimizing these two loss components, the SPINN model learns both from the inherent relationship within data and also satisfies the underlying governing physical relationship dictated by the empirical equation.

In this study, we implement three SPINN variants, with three empirical equations, including the widely accepted HEC-18 equation to a modified time-dependent form as described in the following sections. These equations involve latent parameters calibrated based on time series data throughout the training process. These parameters are optimized by Adam optimizer based on the total loss alongside the hidden parameters of the base NN model \cite{kingma2014adam}.

\subsubsection{SPINN with HEC18 Equation (SPINN-HEC18)}
\noindent
HEC-18, officially known as Hydraulic Engineering Circular No. 18, is a widely recognized engineering guideline published by the United States Federal Highway Administration (FHWA). This guideline provides recommendations for the prediction of maximum depth of scour. Over time, HEC-18 has undergone iterative revisions, with the most current version documented in \citet{Hec-18:2012}. The latest HEC-18 Equation (\ref{eq:2}) is shown in Equation (\ref{eq:2}):

\begin{equation} \label{eq:2}
    y_s=2.0aK_1K_2K_3(\frac{y_1}{a})^{0.35}(\frac{V}{gy_1^{0.5}})^{0.65}
\end{equation}

where $y_s$ is the equilibrium scour depth, $y_1$ is the flow depth, $a$ is the width of the pier, and $K_1$, $K_2$, $K_3$, are correction factors related to the conditions of the pier, the flow and the riverbed as defined in Table~\ref{tab:table_notation}. This equation is predominantly calibrated using datasets from laboratory measurements. 

\begin{figure}
    \centering
    \includegraphics[width=0.95\linewidth]{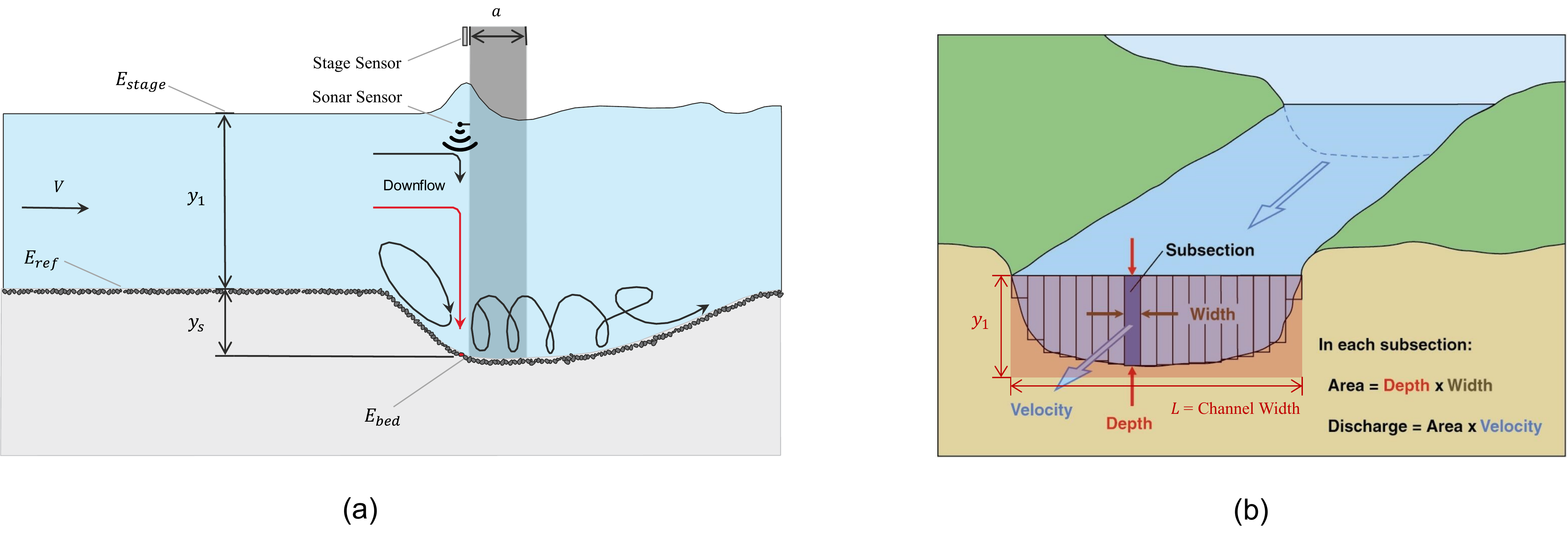}
    \caption{a) Profile of pier subjected to scour, showing sensors and variables [adapted from \protect\citet{Hec-18:2012}], b) Calculation of channel cross-sectional area and the relationship between velocity and discharge.}
    \label{fig:2}
\end{figure}

For training the SPINN model, we use time series of three features collected from continuous sensor measurements at bridge piers, including riverbed elevation, flow elevation and discharge, $E_{bed}$, $E_{stage}$, and $q$. Scour depth and flow depth, $y_s$ and $y_1$ at each timestep can be derived by the following equations, as shown in Fig.~\ref{fig:2}a:

\begin{equation} \label{eq:3}
    y_1=E_{stage}-E_{ref}
\end{equation}
\begin{equation} \label{eq:4}
    y_s=E_{ref}-E_{bed}
\end{equation}

where $E_{stage}$ is the stage (elevation of the flow directly upstream of the pier), $E_{bed}$ is the bed elevation, and $E_{ref}$ is the reference level. For SPINN-HEC18, we use the as-built bed elevation as $E_{ref}$.

We derive average velocity ($V$) based on the time series of discharge data ($q$), as the direct measurements of velocity are not provided by USGS. As shown in Fig.~\ref{fig:2}b, discharge can be estimated by river channel cross-sectional area times velocity. The cross-sectional area of flow ($A$) is not easy to measure nor available in this study, therefore we assume that $A$ can be calculated as $y_{1}\times{L}$ multiplied by a fixed ratio $p_{2}\in\left[0,1\right]$, which can be estimated as a latent parameter within SPINN. Therefore, $V$ can be calculated as:

\begin{equation} \label{eq:5}
V=\frac{q}{A}=\frac{q}{p_2Ly_1}
\end{equation}

In addition, we include $p_1\in\left[-1,1\right]$ as another latent parameter that adjusts the combined correction factors (pier shape, flow direction, and riverbed). Therefore, the empirical equation, ${\hat{Y}}_{s_{PHY}}$ can be written as:

\begin{equation} \label{eq:6}
{\hat{Y}}_{s_{PHY}}=p_{1}2.0K_{1}K_{2}K_{3}\frac{a^{0.65}}{g^{0.215}L^{0.43}y_1^{0.295}}\left(\frac{q}{p_2}\right)^{0.43}
\end{equation}

where $p_1$ and $p_2$ will be calibrated through the SPINN training process, resulting in a new calibrated empirical equation. 

Equation~\ref{eq:6} is only applicable during scouring episodes and not to filling episodes. To account for this, we impose a condition to include the physical loss term in the total loss only when $y_s$ is greater than zero (i.e., $E_{ref}>E_{bed}$). When $y_s$ becomes negative, which happens during filling episodes, the $MSE_{PHY}$ term is set to zero for those specific mini-batches or sequences.

\subsubsection{SPINN with Time-Dependent HEC18 Equation (SPINN-TD)}
\begin{figure}[h]
    \centering
    \includegraphics[width=0.92\linewidth]{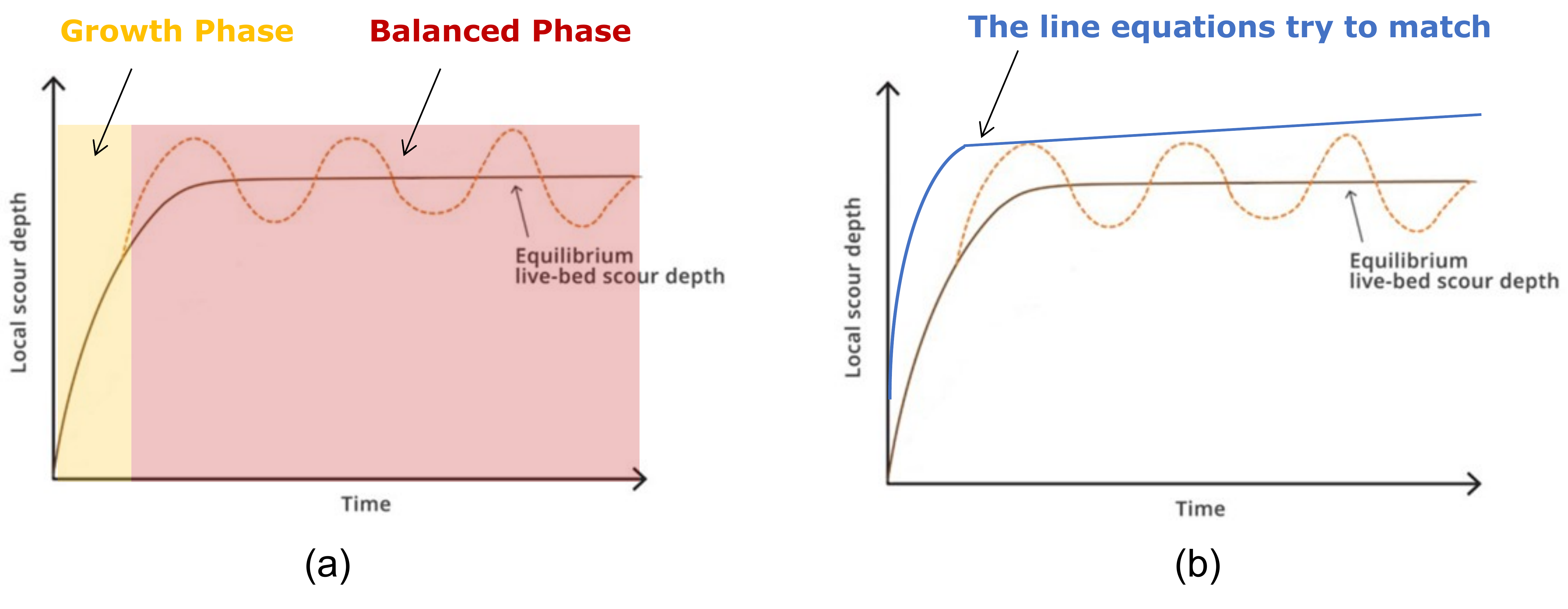}
    \caption{Temporal features of live-bed local scour: a) Different phases of scour evolution, b) The ideal fitting line representing the upper bound of scour depth. Source: \protect\citet{ngo2018guide}.}
    \label{fig:3}
\end{figure}

\noindent
Our dataset records show that the Alaskan bridges experience live-bed scour, where the riverbed goes through iterative cycles of scouring and filling \cite{ngo2018guide}. For each occurrence of a scour event, the progression of live-bed scour can be outlined into two distinct phases, as illustrated in Fig.~\ref{fig:3}a. In the initial growth phase, local scour begins from a depth of zero and gradually deepens. Upon reaching equilibrium depth, the local scour undergoes fluctuations, maintaining a relatively steady state. During this balanced phase, the flow dynamics lead to sediment deposition and erosion, resulting in a state of equilibrium. A number of time-dependent equations \cite{sheppard2011scour,liang2019local} have been developed to approximate the upper bound of scour depth, as depicted in Fig.~\ref{fig:3}b.

\begin{figure}[h]
    \centering
    \includegraphics[width=1.0\linewidth]{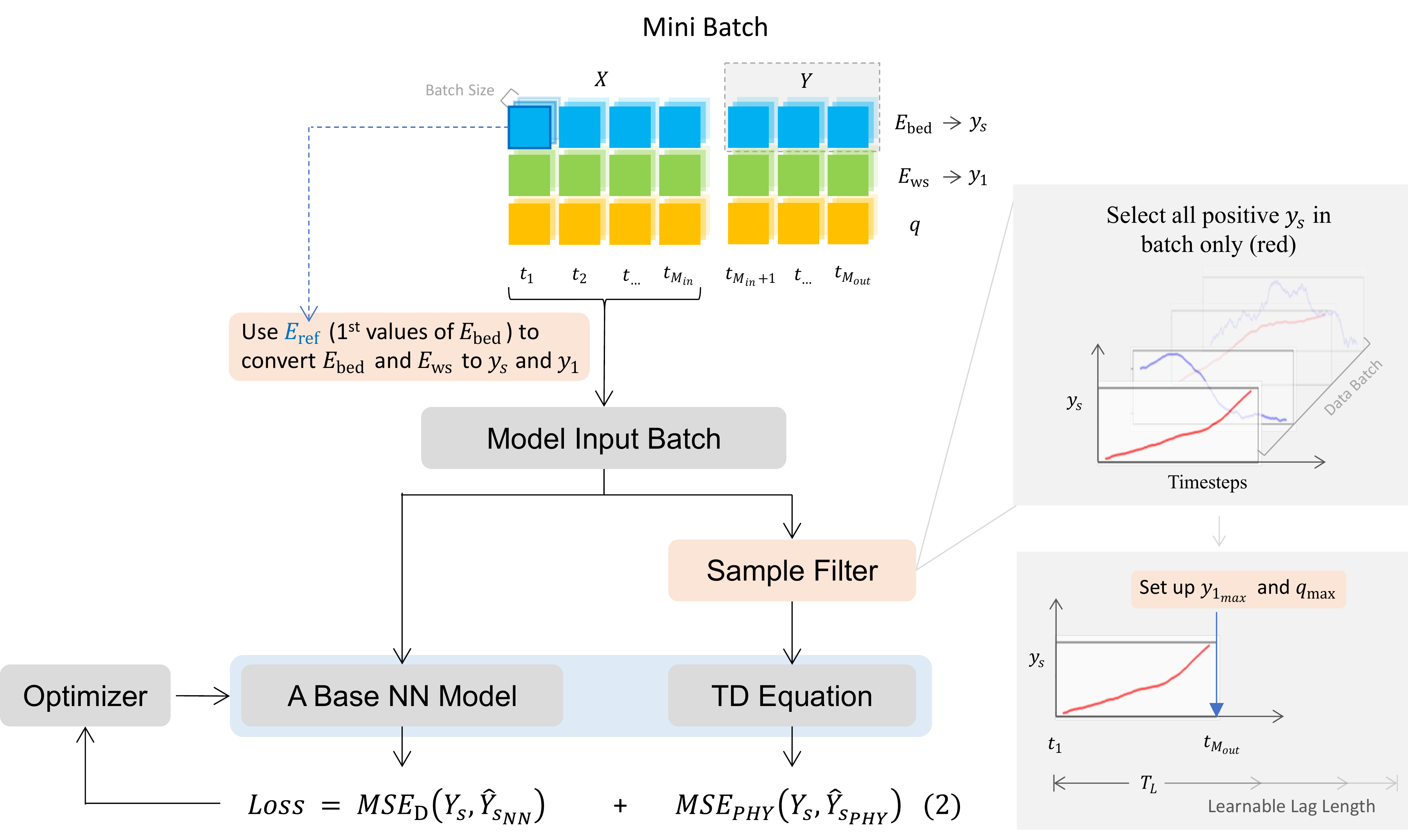}
    \caption{The training process of the time-dependent SPINN model, which incorporates scouring episode detection and applies the physical loss term only during scouring episodes.}
    \label{fig:4}
\end{figure}

We define a time-dependent empirical model of scour based on the maximum scour depth ($y_{s\_max}$) from HEC18 and an exponential decay function to account for time dependency as shown in Equation~\ref{eq:7}. In each data slice within a scour episode, the elevation at the first timestep is considered as $E_{ref}$, and the last timestep $M_{out}$ within the target sequence is considered as the maximum scour depth. 


\begin{equation} \label{eq:7}
    {\hat{y}}_{st}=y_{s_{max}}\ \left(1-e^{-\frac{p3\times t}{T_L}}\right)=p_12.0K_1K_2K_3\frac{a^{0.65}}{g^{0.215}L^{0.43}y_{1_{max}}^{0.295}}\left(\frac{q_{max}}{p_2}\right)^{0.43}\left(1-e^-\frac{p_3\times t}{T_L}\right)
\end{equation}


In this equation, ${\hat{y}}{st}$ is the predicted time-dependent local scour depth at timestep $t$, where $t\in\left\{0,1,2,\ldots,M_{out}-1\right\}$ represents the time steps of the output sequence. $T_L$ is the time lag between the maximum scour depth and the maximum flow, which is often observed in the scour process. This time lag is physically meaningful as it accounts for the delayed response of the scour depth to the flow conditions. $p_1$, $p_2$, and $p_3$ are latent parameters, where $p_3$ represents the decay rate in the exponential function, controlling how quickly the scour depth approaches the equilibrium state. These parameters are passed through a hyperbolic tangent (tanh) activation function, which constrains their values to the range $\left[-1,1\right]$. $T_L$ is set to be within the range $\left[0,\ 2\left(M_{in}+M_{out}\right)\right]$. By setting the upper limit of $T_L$ to be twice the sum of $M_{in}$ and $M_{out}$, we ensure that the parameter has sufficient flexibility to learn the appropriate time lag between the maximum scour depth and the maximum flow, even if the lag extends beyond the output sequence length.

Given that the time-dependent equation mentioned above is specifically applicable to instances of scouring episodes, it is important to identify these episodes as illustrated in Fig.~\ref{fig:4}. The training workflow for a single data batch comprises several key steps. Initially, the bed elevation values of the first timestep within the mini-batch are picked as the reference level $E_{ref}$ for calculating flow depth and scour depth for each sequence, which are inputs for the NN model and the TD empirical equation. During this process, the sequences pass through the conditional term of physics loss ($MSE_{PHY}$) to filter scour episodes, similar to SPINN-HEC18 models. The physics loss is only applied to the scour episodes, therefore $MSE_{PHY}$ term is set to zero for filling episodes/sequences.

Similar to SPINN-HEC18, the training is performed based on total loss for all mini-batches, including both scouring and filling episodes, to adjust the hidden parameters in the NN model and latent parameters in the TD empirical equation.

\subsubsection{Transferable/General Time-Dependent Equation (SPINN-GTD)}
\noindent
Both SPINN-HEC18 and SPINN-TD are site/bridge-specific models and require abundance of site-specific data, including pier geometry, and correction factors for bed condition, and flow attack angle. In order to develop a general model that is transferable across multiple bridges within a region (a cluster of bridges with similarities), we introduce SPINN-GTD by eliminating all site-specific parameters from the SPINN-TD empirical equation (Equation~\ref{eq:7}), as outlined in Equation~\ref{eq:8}:

\begin{equation} \label{eq:8}
{\hat{y}}_{st}=p_1\left(y_{1_{max}}\right)^{-\alpha}\left(q_{max}\right)^{\beta}\left(1-e^-\frac{t}{T_L}\right)
\end{equation}

where $\alpha$ and $\beta$ represent hyperparameters introduced to offer more flexibility in the shape of the empirical model to adjust to a range of bridges. 

This approach enables the SPINN model to get trained with a much larger database and learn the features of a wider range of scour events from a cluster of bridges showing similar flow and scour trends. Such transferable model is envisaged to be able to collate and transfer the learning among several bridges. Transferable models are particularly relevant for new bridges with limited data available. The calibrated general time-dependent (GTD) equation may be viable to provide reasonable estimates of maximum scour depth in a new bridge within the same region/cluster. We evaluated this hypothesis in the following sections.

\subsubsection{Base DL Algorithm: 1-Long Short-Term Memory Networks (LSTM)}
\noindent
LSTM is a type of recurrent neural network (RNN) that excels in capturing and modelling temporal dependencies within time series/sequential data \cite{hochreiter1997long}. A critical challenge in training deep neural networks is the vanishing gradient problem, where gradients shrink to a very small level during back-propagation, which hinders effective learning. LSTM networks address this problem through their unique gate mechanisms, which regulate and maintain the gradient flow, preventing its significant diminish across the network's layers. The strength of LSTMs lies in their ability to remember long-term dependencies in sequential data, allowing them to capture complex patterns that traditional models might miss. This feature is particularly advantageous in time series prediction, where future outcomes are influenced not just by recent events but also by occurrences in the distant past, resulting in more accurate and robust predictions.

\begin{figure}[h]
    \centering
    \includegraphics[width=0.82\linewidth]{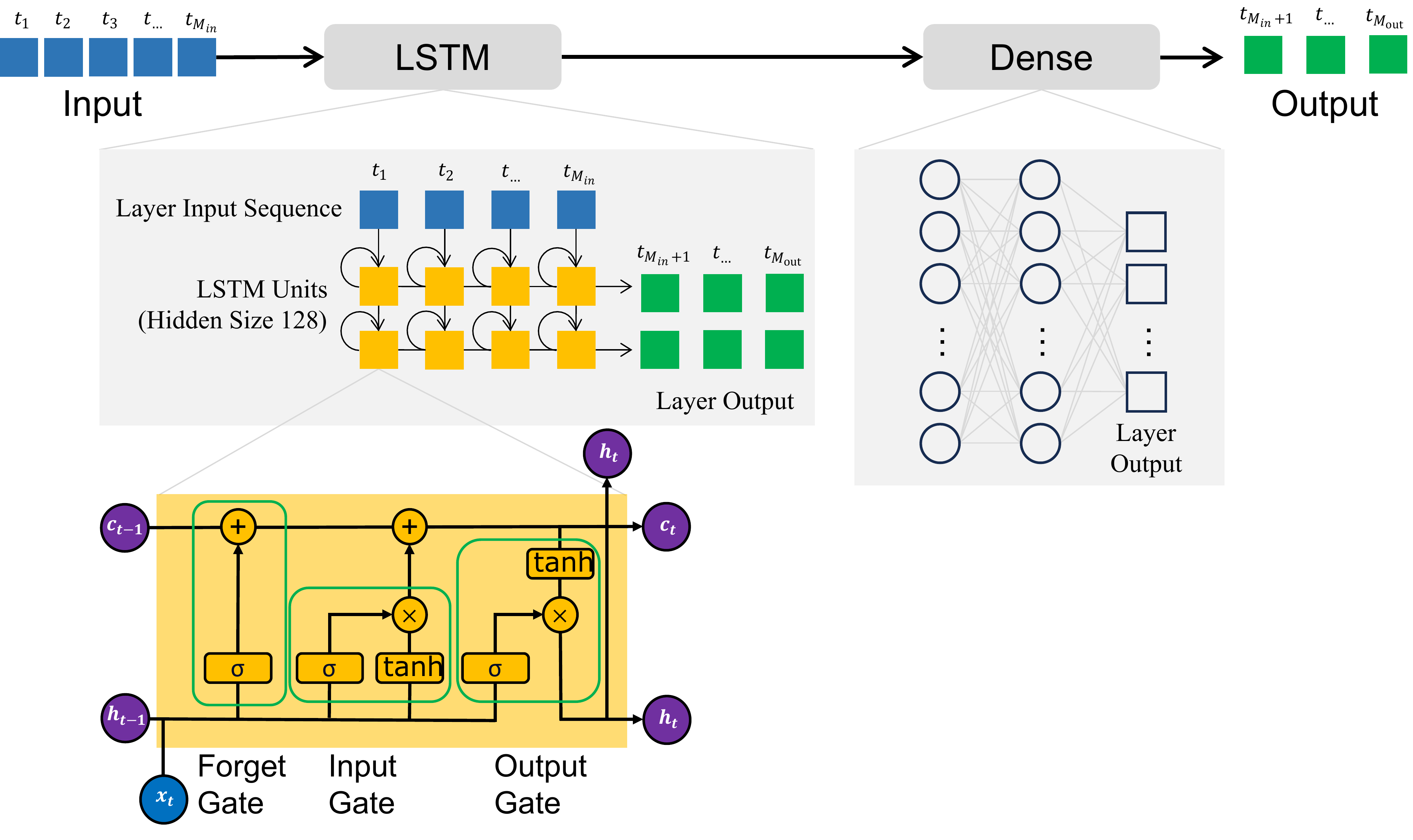}
    \caption{The architecture of LSTM base models for SPINNs, showing the LSTM Memory Unit gates and the input (\( x_t \)), hidden (\( h_t \)), and cell state (\( c_t \)) vectors.}
    \label{fig:model-lstm}
\end{figure}

In this study, we implement a one-layer LSTM model with 128 hidden units as shown in Fig.~\ref{fig:model-lstm}. The input sequence is processed by the LSTM layer, and then passed through a dense (fully connected) layer with a linear activation function and finally reshaped into the output sequence. 

Readers are referred to \citet{yousefpour2023towards} and \citet{hashem2024application} for more details on optimisation and hyperparameter tuning for the  LSTM scour models.

\subsubsection{Base DL Algorithm: 2-Convolutional Neural Network (CNN)}
\noindent
CNN is one of the most prevalent types of neural network applied in computer vision due to its effectiveness in learning image features \cite{venkatesan2017convolutional}. In recent years, they have also found applications in time series prediction tasks, where the inputs are structured as sequential data \cite{livieris2020cnn}. Similar to spatial pattern identification in images, CNN can also be adapted to capture temporal patterns and dependencies within sequential data in the context of time series prediction \cite{liu2022scinet}. 

\begin{figure}[h]
    \centering
    \includegraphics[width=0.85\linewidth]{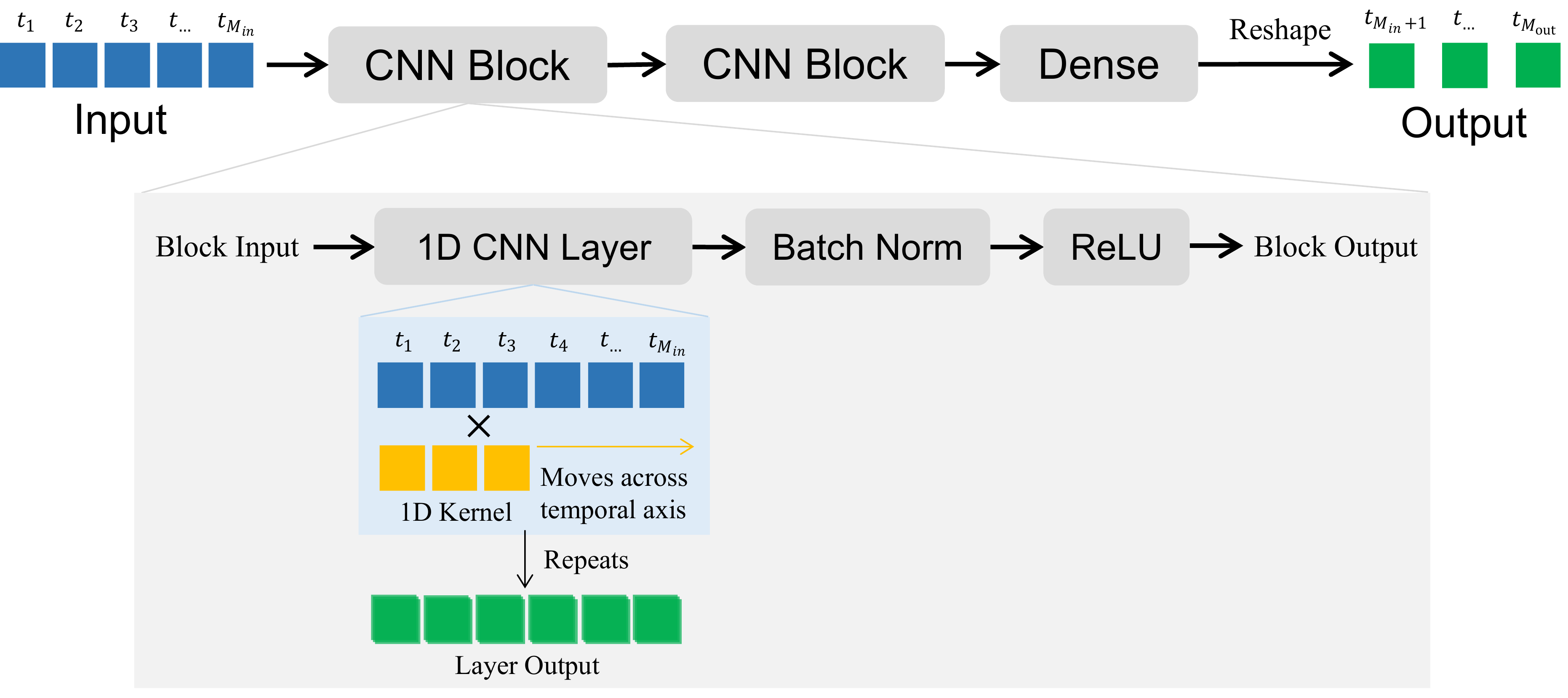}
    \caption{The architecture of the base CNN models for SPINNs.}
    \label{fig:model-cnn}
\end{figure}

In this study, we develop a CNN model with two 2D convolutional layers to show the adaptability of CNNs for processing multidimensional time series data. The two CNN layers have a kernel size of 5, padding size of 2, and output sizes of 128 and 256, respectively. Also, batch normalization layers are added after these layers, and one following dense layer is responsible for producing the final prediction. The CNN model architecture is shown in Fig.~ \ref{fig:model-cnn}.

Readers are referred to \citet{hashem2024application} for details of model optimisation and hyperparameter tuning for the CNN models.

\section{Results and Discussion} \label{sec-results}
\noindent

\subsection{Experimental Design} \label{sec-experimental-design}
\noindent
A number of experiments were performed to evaluate the performance of the three SPINN variants and their counterpart pure NN (base data-driven) models. The experiments can be categorized into general/transferable and site/bridge-specific as shown in Table~\ref{tab:experiment-plan}. 

\begin{table}[h]
\centering
\caption{Experiment design for evaluation of SPINN variants and their corresponding pure data-driven models. The general models trained using a combination of all four datasets indicated as "All". The latent parameters calibrated through the SPINN training process are highlighted.}
\label{tab:experiment-plan}
\begin{tabularx}{\textwidth}{l X X l} 
\toprule
Experiments & Training Dataset & Base Data-Driven Model & $\hat{Y}_{s_{PHY}}$ \\
\midrule
Pure NN (Site-Specific) &
  212\newline 527\newline 539\newline 742&
  LSTM\newline CNN &
  N/A \\
Pure NN (General)&
  All&
  LSTM\newline CNN &
  N/A \\
SPINN-HEC18 (Site-Specific) &
  212\newline 527\newline 539\newline 742 &
  LSTM\newline CNN &
  $\begingroup\color{magenta}p_1\endgroup 2.0K_1K_2K_3\frac{a^{0.65}}{g^{0.215}L^{0.43}y_1^{0.295}}\left(\frac{q}{\begingroup\color{magenta}p_2\endgroup}\right)^{0.43}$  \\
SPINN-TD (Site-Specific) &
  212\newline 527\newline 539\newline 742 &
  LSTM\newline CNN &
  $\begingroup\color{magenta}p_1\endgroup2.0K_1K_2K_3\frac{a^{0.65}}{g^{0.215}L^{0.43}y_{1_{max}}^{0.295}}\left(\frac{q_{max}}{\begingroup\color{magenta}p_2\endgroup}\right)^{0.43}\left(1-e^-\frac{\begingroup\color{magenta}p_3\endgroup\times t}{\begingroup\color{magenta}T_L\endgroup}\right)$  \\
SPINN-GTD  (General) &
  All &
  LSTM\newline CNN &
$\begingroup\color{magenta}p_1\endgroup\left(y_{1_{max}}\right)^{\begingroup-\color{magenta}\alpha\endgroup}\left(q_{max}\right)^{\begingroup\color{magenta}\beta\endgroup}\left(1-e^-\frac{t}{\begingroup\color{magenta}T_L\endgroup}\right)$ \\
\bottomrule
\end{tabularx}
\end{table}

The models trained using bridge-specific data will be evaluated with the corresponding test set, while the model trained on all bridges will be evaluated individually for each of the four bridges. Considering the stochastic nature inherent in NN modelling, each experiment is trained and evaluated five times to ensure the robustness of the results. The averaged MSE values over the validation and test datasets are used to evaluate the performance of models.

In addition to MSE, we also use Mean Absolute Percentage Error (MAPE) and Root Mean Squared Error (RMSE) as performance metrics for evaluating the models. MAPE and RMSE are defined as:

\begin{equation} \label{eq:mape}
MAPE = \frac{1}{M_{out}}\sum_{i=0}^{M_{out}}\left|\frac{y_i-\hat{y}_i}{y_i}\right| \times 100
\end{equation}

\begin{equation} \label{eq:rmse}
RMSE = \sqrt{\frac{1}{M_{out}}\sum_{i=0}^{M_{out}}(y_i-\hat{y}_i)^2}
\end{equation}

where $y_i$ is the actual scour depth and $\hat{y}_i$ is the predicted scour depth at timestep $i$.
 
The models were implemented using PyTorch \cite{paszke2017automatic} python packages and were run on 4 NVIDIA A100 Tensor Core GPUs and 2$\times$16 Core CPUs on the \textit{Geo\&Co} Infrastructure Department High-Performance Computing Center at the University of Melbourne.

\subsection{DL Model Configuration and Training}
\noindent
Fig.~\ref{fig:train_val_test_539} shows the historical time series data of Bridge 539 as an example (see Appendix Section for time series data for other case study bridges). Multiple data gaps are found within the data, which is mainly seasonal freezing in Alaska. We use a sliding window with a length equal to $M_{in}$ plus $M_{out}$ to slice the data into the sequences for training. 

Based on the top model configurations identified through comprehensive hyperparameter tuning performed by \citet{yousefpour2023towards} and \citet{hashem2024application}, the input width, $M_{in}$ and output/label width, $M_{out}$ for all models are set to 168 hours (seven days). This means that after training the models with the historical time series data up to the current time, the last one-week data is used as input to predict the scour depth for the coming week.

\begin{figure}[htb!]
    \centering
    \includegraphics[width=1.0\linewidth]{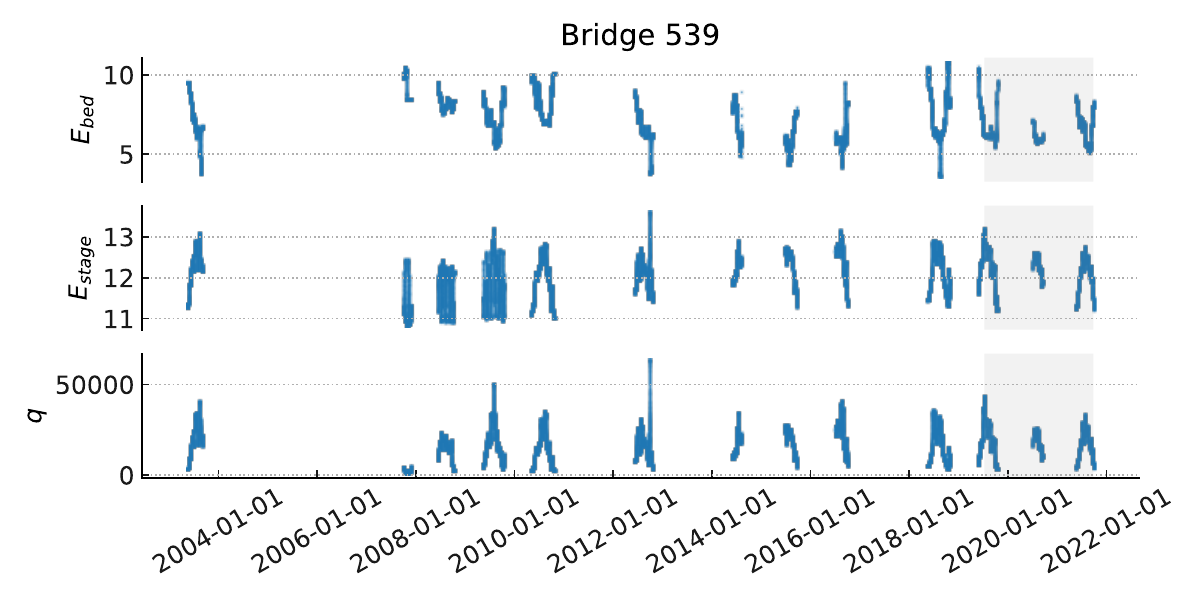}
    \caption{
    Historical time series data for Bridge 539 shows the three features: bed elevation ($m$), river stage elevation ($m$), and discharge ($m^3/s$). The grey zone shows the test subset, and the transparent part before is the training and validation subsets.}
    \label{fig:train_val_test_539}
\end{figure}

\begin{table}[h]
\centering
\caption{The split of training, validation, and testing sets for each bridge.}
\label{tab:train_val_test_split}
\begin{tabular}{ccllc}
\toprule
Bridges & Type & Start & End & No. of Sequences \\
\midrule
212 & Train/Validation & 2012-06-05 13:00 & 2017-08-06 02:00 & 11,573 \\
212 & Test & 2017-08-06 03:00 & 2018-09-27 03:00 & 2725 \\
527 & Train/Validation & 2005-06-02 15:00 & 2019-06-11 11:00 & 41,473 \\
527 & Test & 2019-06-11 12:00 & 2021-09-25 22:00 & 9782 \\
539 & Train/Validation & 2003-05-13 17:00 & 2019-07-10 17:00 & 29,626 \\
539 & Test & 2019-07-10 18:00 & 2021-09-25 20:00 & 7155 \\
742 & Train/Validation & 2013-05-27 02:00 & 2017-07-09 23:00 & 7,462 \\
742 & Test & 2017-07-10 00:00 & 2017-09-11 17:00 & 1,530 \\
\bottomrule
\end{tabular}
\end{table}

For each bridge, we have time series data across several years. We create input-output pairs by sliding a window over the continuous time series data, resulting in a number of data slices. We arrange these slices in temporal order and allocate the last 20\% for the test set. The first 80\% of the samples are then randomly split into training and validation sets in a 3:1 ratio (60\% for training and 20\% for validation). The model is validated after each training epoch to monitor its performance on unseen data and prevent overfitting. The range of available time series data and the number of available sequences over training/validation and test datasets in each bridge is provided in Table~\ref{tab:train_val_test_split}. Fig.~\ref{fig:train_val_test_539} illustrates the data split for Bridge 539. Similar illustrations for other bridges can be found in the Appendix Section (Figs. \ref{fig:train_val_test_212} to \ref{fig:train_val_test_742}).

\subsection{DL Model Performance Variation Across Case Studies}
\noindent

\begin{figure}[h!]
    \centering
    \includegraphics[width=1\linewidth]{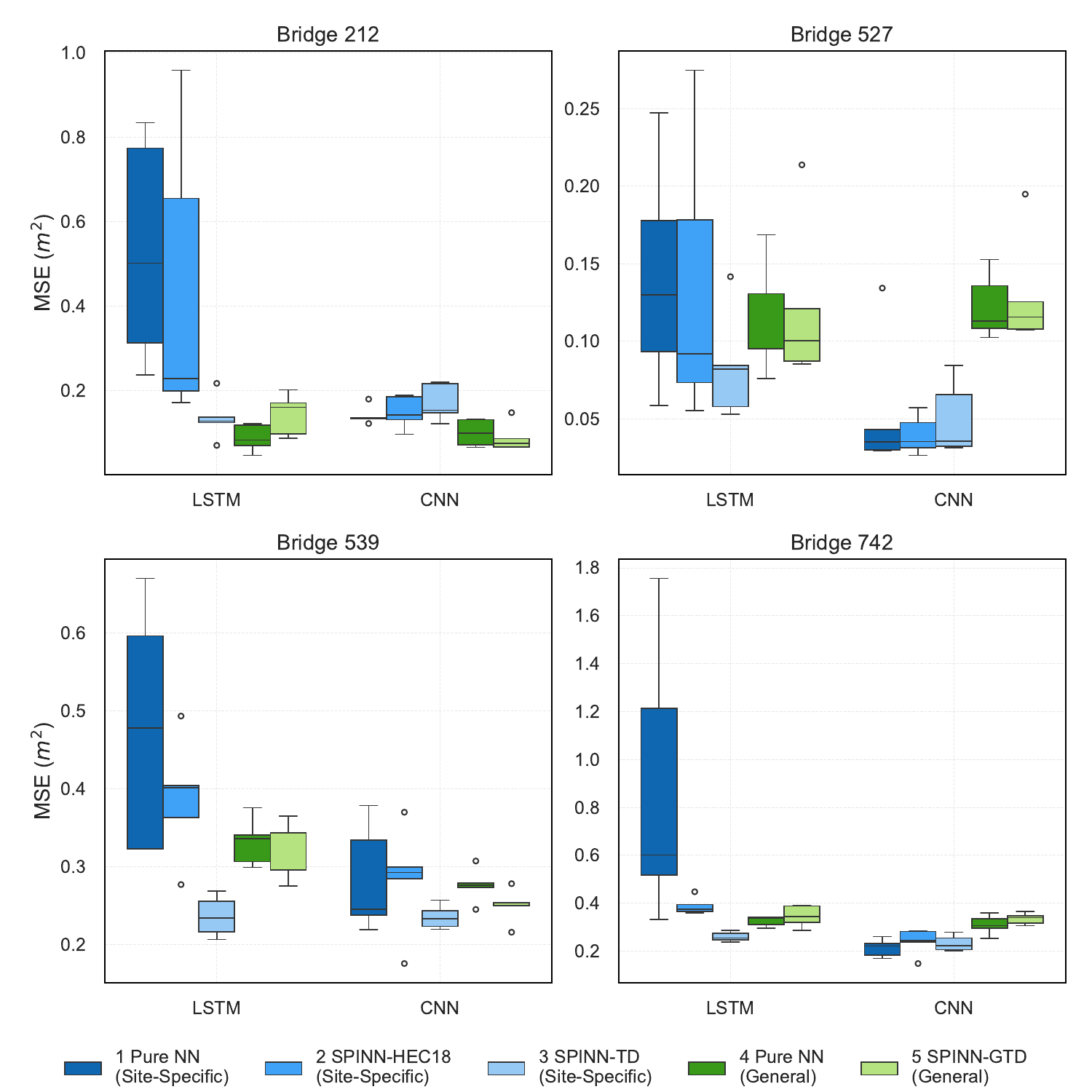}
    \caption{Model performance across four case study bridges, showing the test MSE ($m^2$) boxplots for site-specific (blue) and general (green) Pure NN and SPINN variants.}
    \label{fig:result_TestMSE}
\end{figure}

Fig.~\ref{fig:result_TestMSE} presents the boxplots of performance (MSE) for all models over the test dataset. The detailed performance of all variants is provided in Table~\ref{tab:table_overall_performance}. The aggregated MSE of five runs is listed as the mean and standard deviation values for each model. 

Table~\ref{tab:performance-pure} presents the MAPE and MSE criteria for pure data-driven models across case study bridges. It is shown that the order of performance across four bridges is different for the two performance criteria. MAPE is a percentage-based metric that provides insights into the relative magnitude of errors and is particularly useful when comparing models across bridges with varying scales of bed elevations. On the other hand, MSE concentrates larger errors by squaring them, making it more sensitive to outliers. Also, the trend in model performance across different bridges does not show a correlation with the amount of training data available for each bridge (see Table~\ref{tab:train_val_test_split}). 

The DL model performance for each bridge can be influenced by data quality, scale of variation, amount of missing data, and consistency in seasonal and temporal patterns across years.
As observed in Fig.~\ref{fig:train_val_test_539}, and also Fig.~\ref{fig:train_val_test_212} to \ref{fig:train_val_test_742}, the degree of fluctuations in the bed elevation, $E_{bed}$, varies across case study bridges. For instance, Bridge 539 experienced several significant scour events across the years, with some scour depths reaching over five meters. In contrast, bed elevation changes at Bridge 527 were small, with variations consistently below two meters. Similarly, bridge 212 does not show significant scour and filling throughout the years. It should be noted that this variability in scale and rate of variation in scour depth can affect the relative performance of different DL algorithms. This complexity should be acknowledged in comparing and assessment of competing models.

\begin{table}[h]
\centering
\small
\caption{Performance of site-specific and general pure data-driven models for four case study bridges, showing MSE ($m^2$) / MAPE ($\%$) values. Elev. represents the maximum bed elevation variation ($m$) over the test dataset.}
\label{tab:performance-pure}
\begin{tabular}{lllllllllll}
    \toprule
     & \multicolumn{1}{l}{} & \multicolumn{2}{c}{Pure-LSTM} & \multicolumn{2}{c}{Pure-CNN} \\
 & \multicolumn{1}{l} {Elev.} & Site-Specific    &  General   & Site-Specific    & General   \\
 \midrule
212 & 0.644 & 0.533/0.069 & 0.089/0.034 & 0.142/0.041 & 0.101/0.033 \\
527 & 2.030  & 0.141/0.010 & 0.113/0.010 & 0.055/0.006 & 0.123/0.010\\
539 & 4.603 & 0.478/0.537 & 0.332/0.546 & 0.283/0.406 & 0.276/0.485 \\
742 & 3.191 & 0.884/0.127 & 0.325/0.111 & 0.214/0.079 & 0.310/0.106 \\
\bottomrule
\end{tabular}
\end{table}


\subsection {Assessment of SPINN Variants}
\noindent
SPINNs show improvement in predictive accuracy compared with pure NN models in majority of cases. As observed from Fig.~\ref {fig:result_TestMSE}, SPINNs help with reducing mean error and/or reducing the standard deviation of error for 9 out of 16 experiments (4 bridges, 2 base models, site-specific and general). Refer to Table \ref {tab:table_overall_performance}
for a summary of performance (MSE) from all cases/experiments. Table \ref{tab:scour_filling_events_all} provides mean and standard deviation of MSE for SPINN variants versus their corresponding pure NN models for bridge 539 and 742. 

For the site-specific category, the SPINN variants showed considerably better performance compared to their pure NN counterparts for the LSTM group. In the case of Bridge 539 for instance, the SPINN-TD variant reduces the MSE by 50\% (from 0.48 to 0.24$m^2$). The SPINN-HEC18 variant also enhances the performance of LSTM by 19\% (from 0.48 to 0.39 $m^2$). Similarly, for Bridge 742, the SPINN-TD variant reduces the error by 70\% (MSE from 0.88 to 0.26$m^2$) for the LSTM base model. In CNN group, for bridge 539, SPINN-TD reduces the MSE by 17\% (from 0.28 to 0.24$m^2$), however, for the other three bridges SPINN variants and pure CNN models show close performance with MSE average less than 0.3$m^2$.

For the general/transferable models, SPINN-GTD shows comparable performance with marginal improvement. The best outcome is observed for the CNN model of bridge 539 with 9\% (from 0.276 to 0.250) MSE reduction. 


\begin{table}[h]
    \centering
    \caption{Performance of SPINN variants and Pure NN models, showing MSE ($m^2$) in bed elevation prediction over test datasets for Bridges 539 and 742.}
    \label{tab:scour_filling_events_all}
    \begin{tabular}{lllll}
    \toprule
    Model   & Training Datasets & Method      & MSE - 539      & MSE - 742      \\
    \midrule
    LSTM    & Site Specific     & Pure NN     & 0.478±0.157     & 0.884±0.589     \\
    LSTM    & Site Specific     & SPINN-HEC18 & 0.388±0.078     & 0.388±0.036     \\
    LSTM    & Site Specific     & SPINN-TD    & 0.236±0.026     & 0.260±0.021     \\
    LSTM    & General           & Pure NN     & 0.332±0.031     & 0.325±0.021     \\
    LSTM    & General           & SPINN-GTD   & 0.325±0.037     & 0.346±0.045    \\
    CNN     & Site Specific     & Pure NN     & 0.283±0.069     & 0.214±0.037     \\
    CNN     & Site Specific     & SPINN-HEC18 & 0.285±0.070     & 0.239±0.055     \\
    CNN     & Site Specific     & SPINN-TD    & 0.236±0.015  & 0.233±0.033     \\
    CNN     & General           & Pure NN     & 0.276±0.022     & 0.310±0.040     \\
    CNN     & General           & SPINN-GTD   & 0.250±0.022     & 0.335±0.024    \\
    \bottomrule
    \end{tabular}
\end{table}

\begin{table}[h]
\centering
\small
\caption{Comparison of SPINN variants across four case study bridges, based on MSE ($m^2$) values. Elev. represents the maximum bed elevation variation ($m$) over the test dataset.}
\label{tab:performance-spinn}
\begin{tabular}{clllllll}
\toprule
\multicolumn{1}{l}{} & & \multicolumn{2}{c}{SPINN-HEC18} & \multicolumn{2}{c}{SPINN-TD} & \multicolumn{2}{c}{SPINN-GTD} \\
\multicolumn{1}{l}{} & Elev. & LSTM & CNN & LSTM & CNN & LSTM & CNN \\
\midrule
212 & 0.644 & 0.443 & 0.150 & 0.136 & 0.173 & 0.144 & 0.089 \\
527 & 2.030 & 0.135 & 0.040 & 0.084 & 0.050 & 0.122 & 0.130 \\
539 & 4.603 & 0.388 & 0.285 & 0.236 & 0.236 & 0.325 & 0.250 \\
742 & 3.191 & 0.388 & 0.239 & 0.260 & 0.233 & 0.346 & 0.335 \\
\bottomrule
\end{tabular}
\end{table}

Table~\ref{tab:performance-spinn} compares the performance of different SPINN variants across the four case study bridges. The SPINN-TD model consistently outperforms the other SPINN variants, achieving the lowest MSE for most of the bridges when paired with LSTM and CNN base models. This suggests that incorporating the proposed site-specific time-dependent equation (TD) as a physics-based loss function is more effective compared to the site-specific HEC18 and the generalized version of the TD equation (GTD). However, the degree of performance improvement varies depending on the specific bridge and base NN model. For instance, while SPINN-TD shows significant error reduction for Bridge 539 when paired with LSTM (50\% reduction) and CNN (17\% reduction), the improvement is less pronounced for Bridge 527 with the same base models (40\% and 9\% reduction, respectively). This variability in performance gain suggests that the effectiveness of the physics-based loss functions is influenced by the inherent characteristics of each bridge dataset.

\subsection{Site-Specific vs General/Transferable DL Models} \label{sec-result-sitespec_and_pureDDM}
\noindent

The general pure data-driven models (Pure NN) show better performance (lower MSE and lower standard deviation) compared to site-specific models, except for CNN in the case of Bridge 212 and 539 (see Table~\ref{tab:performance-pure} and Fig.~\ref{fig:result_TestMSE}). For instance, training the LSTM model with the combined datasets from all bridges led to $\sim$60\% and $\sim$80\% reduction in MSE for bridges 742 and 212, respectively. It is noted that these two bridges had less amount of site-specific data. This indicates aggregating data from a cluster of similar bridges can potentially enhance the generalization of the DL models. This outcome highlights the potential for developing transferable scour forecast models applicable to a cluster of bridges in a region. 

However, the general hybrid models, that is, SPINN-GTD did not show consistent superiority over site-specific hybrid models, i.e. SPINN-HEC18 or SPINN-TD (except for 539 LSTM, 742 LSTM, 212 CNN cases). The challenge of transferability in the SPINN-GTD model is that the general empirical equation is being fit to a wider range of scour characteristics. The GTD equation in fact combines all the site/bridge-specific coefficients and inputs of the site-specific TD equation (i.e., riverbed, flow and pier shape/dimension factors) into one latent parameter ($p_1$). This could limit the extent to which the empirical model can help NN learn and adapt adequately to the inherent scour patterns for each bridge, thus reducing the prediction accuracy. 

\subsection{Evaluation of Base DL Algorithms for SPINNs}
\noindent
As shown in Table~\ref{tab:performance-pure} the relative performance of pure DL models varies across bridges, however, CNN seems to overall outperform LSTM in prediction accuracy, however not in computational cost as shown in Table~\ref{tab:parameters-flops}. 

The comparison of computational cost between LSTM and CNN is expressed in terms of Floating Point Operations (FLOPs). In deep learning, FLOPs are estimated by counting the total number of multiplication and addition operations required for one forward pass through the model. The number of trainable parameters reflects the model's capacity to learn from data, while FLOPs demonstrate the computational demand for training. 

\begin{table}[h]
\centering
\caption{Latent Parameters and FLOPs for Different Models.}
\label{tab:parameters-flops}
\begin{tabular}{lll}
\toprule
Model        & Trainable Parameters & FLOPs       \\
\midrule
LSTM         & 3,681,448            & 22,966,272  \\
CNN          & 7,393,704            & 42,513,408  \\
\bottomrule
\end{tabular}
\end{table}

As shown in Table~\ref{tab:parameters-flops}, there is a direct correlation between the model's complexity and its computational cost. The CNN architecture requires approximately twice the computational effort of the LSTM due to its complex convolutional layers. Regarding the SPINN framework, HEC18, TD, and GTD only introduce a few latent parameters to the base NN (See Table~\ref{tab:experiment-plan}), thereby leading to a negligible extra computational cost. This enables SPINN variants to leverage the benefits of physics-inspired learning without incurring a substantial computational burden.

LSTM shows to achieve significant performance improvement in terms of accuracy (error) and robustness (variance) from a physics-inspired architecture. Given the variability of scour data in various bridges, it is important to investigate competing NN base models to achieve the most optimal performance.

\begin{figure}
    \centering
    \includegraphics[width=0.9\linewidth]{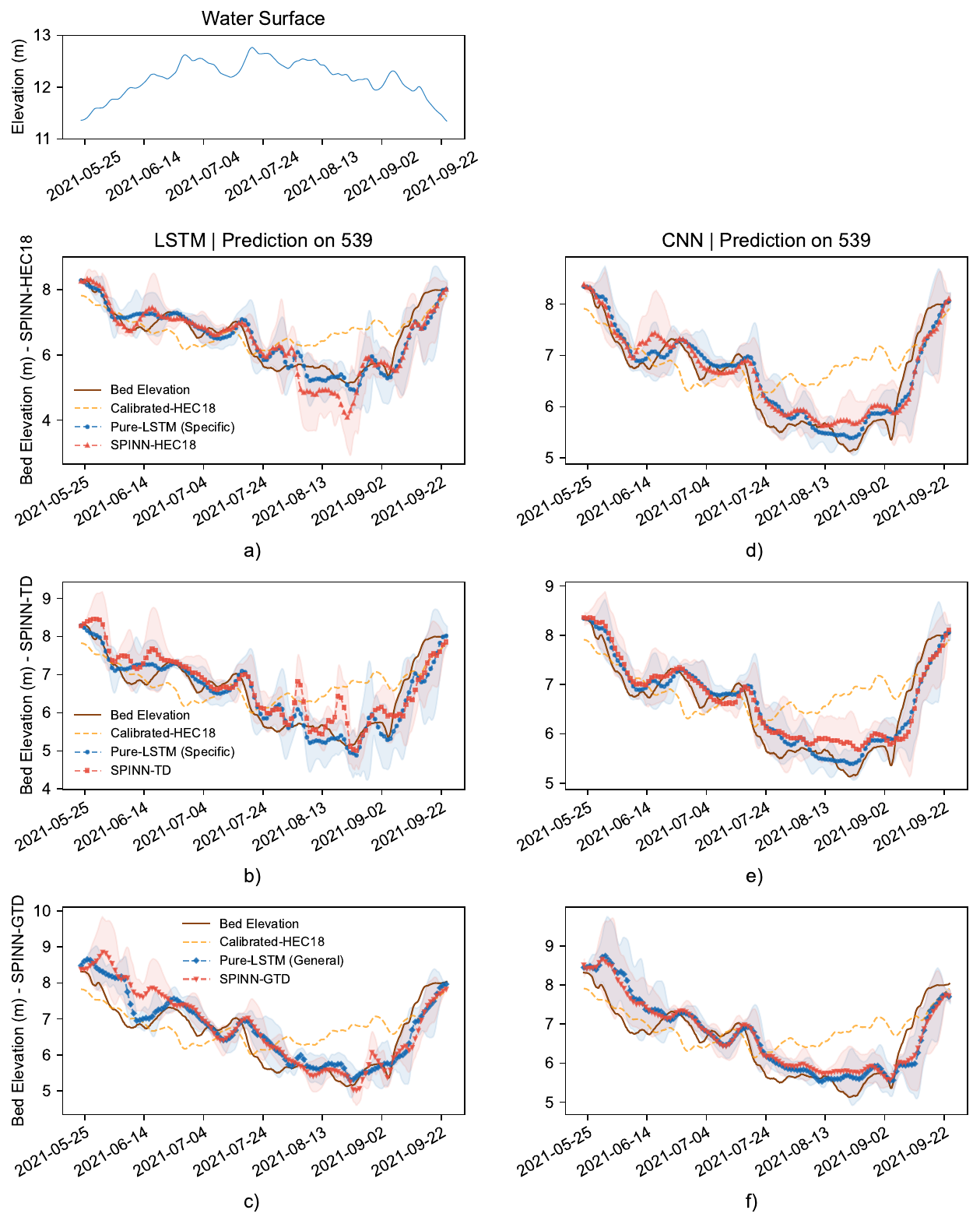}
    \caption{Model predictions versus actual observations for Bridge 539 test set (May-September 2021). Top graph displays the flow changes in this period. (a-c) show LSTM-based predictions, (d-f) display CNN-based predictions, and each row represents SPINN-HEC18, SPINN-TD, and SPINN-GTD, respectively. Additional test periods are presented in the Appendix.}
    \label{fig:7}
\end{figure}

\begin{figure}
    \centering
    \includegraphics[width=0.9\linewidth]{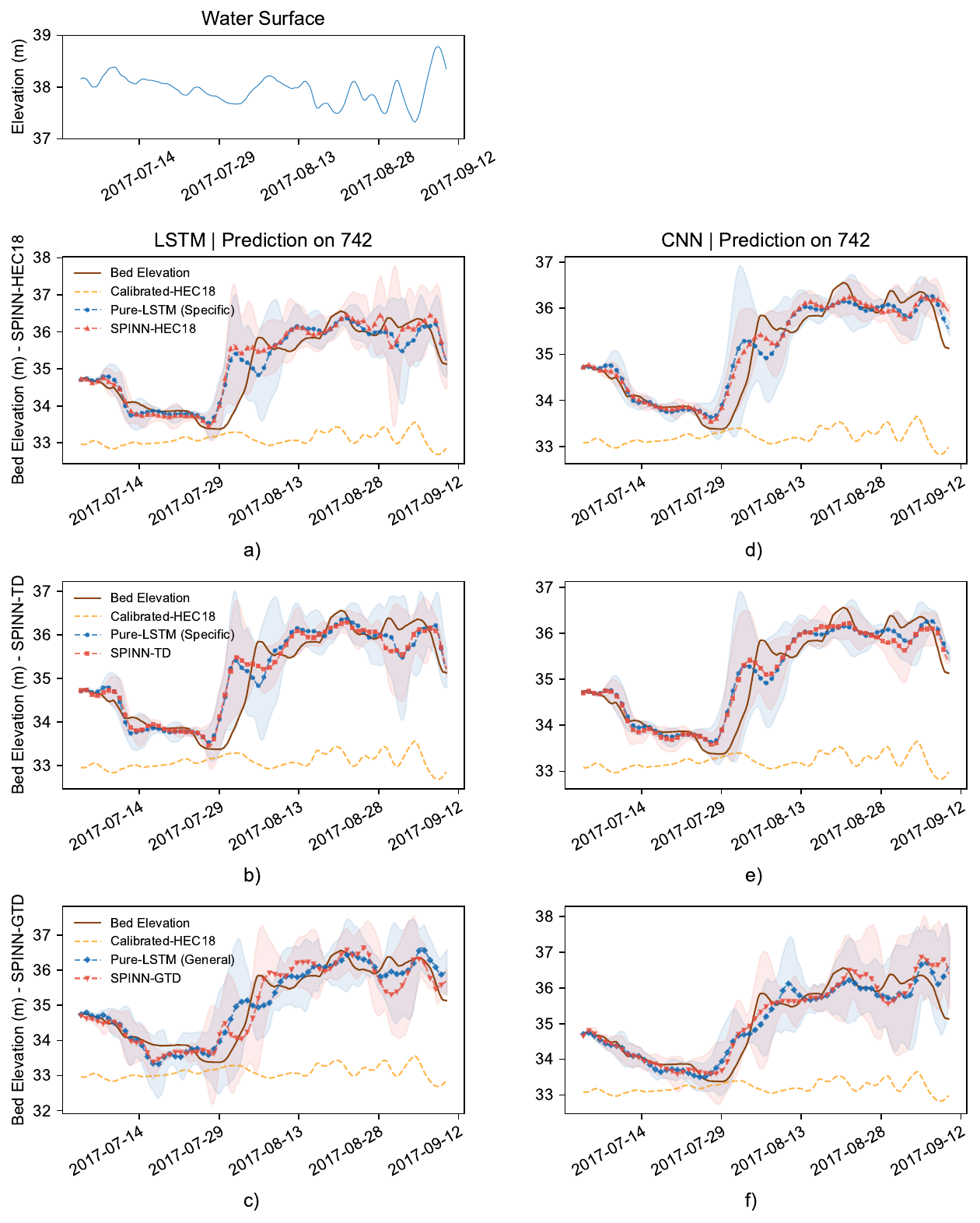}
    \caption{Model predictions versus actual observations for Bridge 742 test set (July-September 2017). Top graph displays the flow changes in this period. (a-c) show LSTM-based predictions, (d-f) display CNN-based predictions, and each row represents SPINN-HEC18, SPINN-TD, and SPINN-GTD, respectively.}
    \label{fig:8}
\end{figure}

\subsection{Forecasting Trends and Capturing Scour Events}
\noindent
The primary objective of the scour forecasting model lies in its ability to predict major scour and filling events in advance. Fig.~\ref{fig:7} and Fig.~\ref{fig:8} present the SPINN model predictions over the test dataset for the three SPINN models. Due to relatively stable bed elevation in the test sets of Bridge 212 and 527, our analysis here is focused on Bridges 539 and 742. The actual river bed elevation changes are depicted as a solid brown line, while the model predictions are shown as dashed lines. The predictions using pure NN models and the calibrated HEC18 empirical equation (CEE-SPINN-HEC18) derived from SPINN (see Section~\ref{sec-calibrated-empiricale-qquations}) are used as baselines for comparison (see also Table~\ref{tab:scour_filling_events_all}).

Comparing both data-driven and hybrid-physics-data-driven models with the calibrated HEC18 equation (CEE-SPINN-HEC18) shows the superiority of the DL models for scour prediction compared with an empirical model. The calibrated HEC-18 model struggles to follow the scouring and filling trends and significantly underestimates the scour depth in most periods for bridges 539 and 742 (see Figs. \ref{fig:7} and \ref{fig:8}).


\subsection{The SPINN-Calibrated Empirical Equations}\label{sec-calibrated-empiricale-qquations}
\noindent
In this section, we shift the focus towards the calibrated empirical equations derived from the proposed SPINN framework. We will evaluate the performance of the three calibrated equations (see Table~\ref{tab:experiment-plan}) and assess their potential to be implemented as independent scour models.

Since the empirical equations are designed to predict maximum scour depth, their performance is only evaluated during scouring episodes. Bridge 539 paired with the LSTM group is chosen here as an example for detailed analysis, considering the deeper scouring depth and the superior performance of the base model.

\begin{figure}[!ht]
    \centering
    \includegraphics[width=0.95\linewidth]{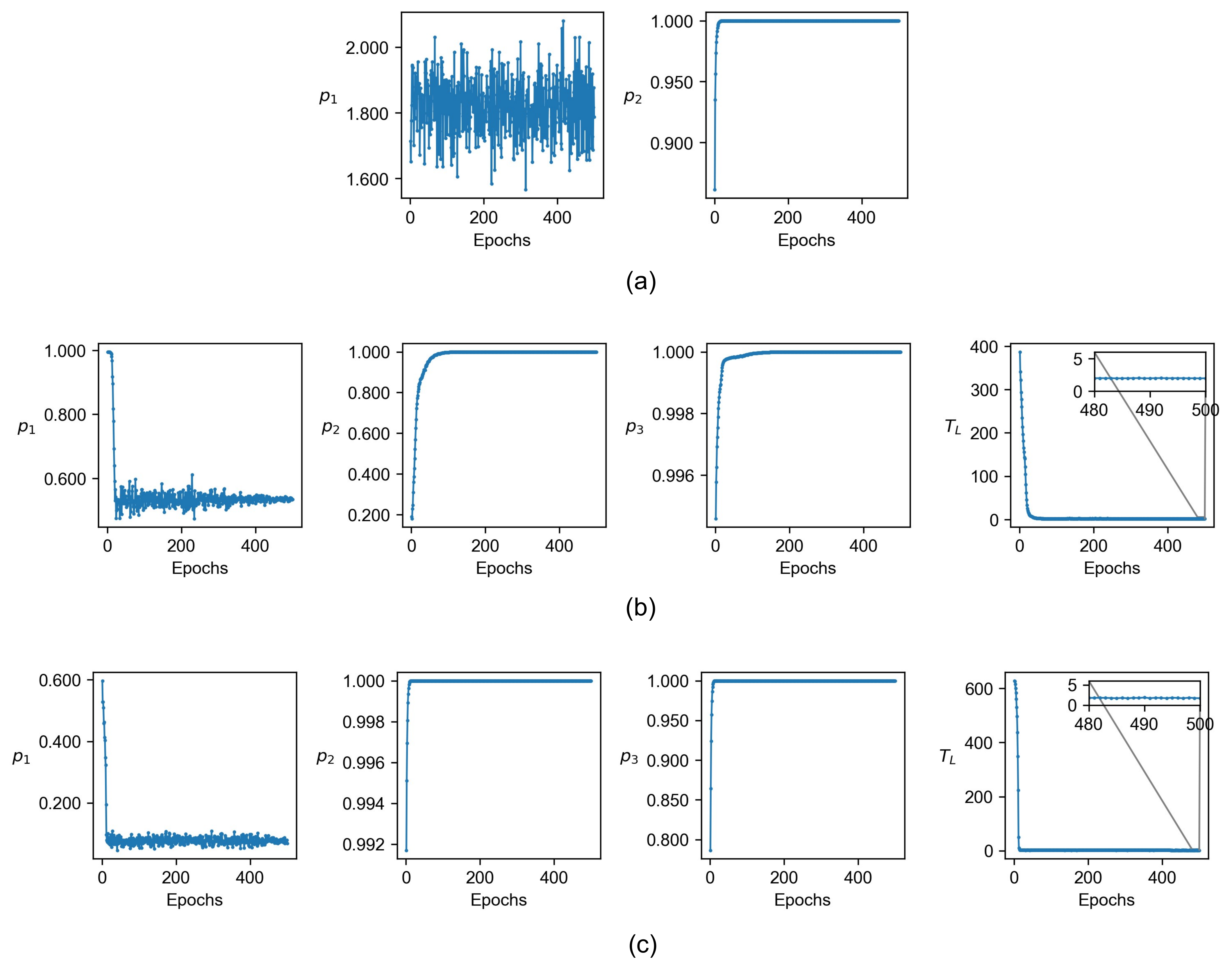}
    \caption{The calibration of empirical equations with LSTM SPINN model, using the 539 dataset for site-specific models (GTD uses all datasets). a) The latent parameters of the HEC18 equation, b) The latent parameters of the TD equation, and c) The latent parameters of the GTD equation.}
    \label{fig:9}
\end{figure}

Fig.~\ref{fig:9} displays the training progress plots of the latent parameters incorporated in the three empirical equations over 500 training epochs. The parameter $p_1$ in all equations acts as an overall adjustment factor, converging to distinct values for each equation, indicating its role in tailoring the output of each empirical equation. The $p_2$ in HEC18 and TD, representing the adjustment on the velocity, stabilizes at 1.0 after initial training epochs, suggesting its impact may be mitigated by adjustments in $p_1$. Similarly, $p_3$ in TD converges to 1.0, reflecting its minimal influence on the final results. $T_L$ is calibrated to various values for site-specific models and to 2.3 hrs for the general model. The summary of calibrated parameters across different scenarios can be found in Table~\ref{tab:equation_params}.
 
\begin{table}[!h]
\centering
\caption{The calibrated equation parameters from SPINN models.}
\label{tab:equation_params}
\begin{tabular}{llllllll}
\toprule
Bridge & Equation & $p_1$ & $p_2$ & $p_3$ & $T_L$ (hour) & $\alpha$ & $\beta$ \\
\midrule
212 & HEC18 & 2.480  & 1.000 & N/A & N/A   & N/A   & N/A   \\
212 & TD    & 0.534  & 1.000 & 1.000 & 1.970  & N/A   & N/A   \\
527 & HEC18 & 0.371  & 1.000 & N/A & N/A   & N/A   & N/A   \\
527 & TD    & 0.098  & 1.000 & 1.000 & 0.356 & N/A   & N/A   \\
539 & HEC18 & 1.790  & 1.000 & N/A & N/A   & N/A   & N/A   \\
539 & TD    & 0.514  & 1.000 & 1.000 & 1.750  & N/A   & N/A   \\
742 & HEC18 & 1.350  & 1.000 & N/A & N/A   & N/A   & N/A   \\
742 & TD    & 0.329  & 1.000 & 1.000 & 9.470  & N/A   & N/A   \\
All & GTD   & 0.076  & N/A & N/A & 2.290  & 0.903 & 0.592 \\
\bottomrule
\end{tabular}
\end{table}

\begin{figure}[!h]
    \centering
    \includegraphics[width=0.45\linewidth]{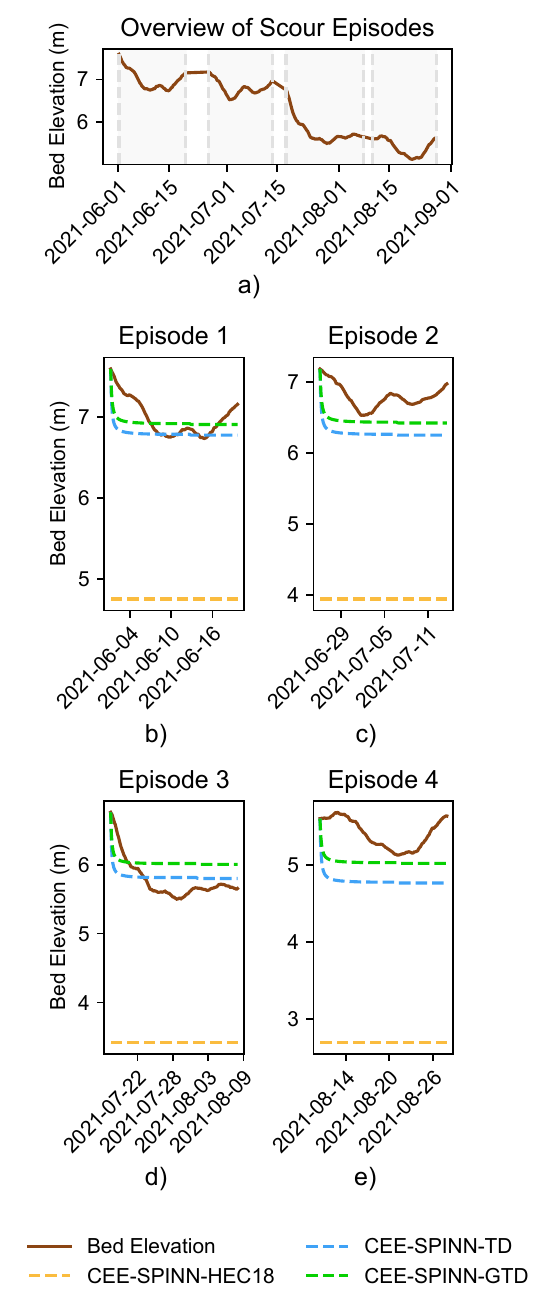}
    \caption{Maximum scour prediction using the calibrated empirical models (CEE) for Bridge 539, a) the overview of bed elevation changes; (b-e) prediction on each scour episode using three SPINN-calibrated equations.}
    \label{fig:calibrated-empirical-equations}
\end{figure}

Fig.~\ref{fig:calibrated-empirical-equations} compares the calibrated empirical equations (CEE) derived from the SPINN model for predicting maximum scour depth on the Bridge 539 test dataset. The actual bed elevation changes are highlighted with a solid brown line, while the predictions from the CEEs are marked with dashed lines. The CEE-SPINN-TD and CEE-SPINN-GTD employ time-dependent equations, where an exponential function predicts the scour depth for each timestep in the scouring episode. The predicted depth is zero at the beginning of the episode ($t=0$) and reaches the maximum scour depth ($y_{s_{max}}$) at a time close to the calibrated time lag parameter ($T_L$). On the other hand, the CEE-SPINN-HEC18 uses the same dynamic reference level approach as the time-dependent equations to estimate the maximum scour depth of each episode (horizontal dashed line).

Among the CEEs calibrated through SPINN, the HEC18 overestimates the maximum scour depth by up to 3 meters. Across these scouring episodes in Fig.~\ref{fig:calibrated-empirical-equations}, the CEE-SPINN-HEC18, CEE-SPINN-TD, and CEE-SPINN-GTD yield RMSEs ($m$) of 2.539, 0.443, and 0.34 respectively. Although not as accurate as SPINN models, the proposed SPINN-calibrated time-dependent empirical equations show great potential to replace existing empirical models for maximum scour depth estimation.

\section{Conclusion}
\noindent
This paper introduced scour physics-inspired neural network algorithms, SPINNs, exploring hybrid physics-data-driven models for bridge scour prediction based on deep learning. Central to SPINN's methodology is the integration of physics-based empirical equations as supplementary loss components into deep neural networks. Although SPINNs performance showed variability with the choice of base DL model, the empirical equation and the case study bridge, they showed superior performance in the majority of scenarios compared to pure data-driven (NN) models. The bridge-specific variants of SPINN with the proposed time-dependent (TD) empirical equation showed the most competitive performance among other SPINN variants. For some scenarios, such as the LSTM group for Bridge 539, the SPINN could help the base NN model reduce forecasting error by up to 50\%.
 
We explored the idea of transferable models for bridge clusters by aggregating bridge datasets across a cluster to train general DL models that can provide reasonable scour prediction for any bridge within that cluster. The general pure data-driven (NN) models showed superiority over site-specific models, however, the general SPINN with the proposed empirical model did not provide as accurate predictions as site-specific SPINNs. 

This study revealed that the sister calibrated empirical equations derived from SPINN framework, particularly the introduced time-dependent equation, can provide a viable alternative to traditional scour models, such as FHWA HEC-18. As opposed to traditional HEC-18, which is often calibrated using discrete scour lab measurements, the empirical equations within SPINN are calibrated using historical time series of scour monitoring data throughout the neural network training process. The proposed general time-dependent calibrated equation can be particularly helpful in scenarios where site-specific flow and riverbed conditions are not readily available (e.g., in new bridge projects).

Comparing SPINNs and the SPINN-calibrated empirical equations with traditional empirical models presents a significant improvement, which can benefit the current state of practice in bridge scour design and management. This paper can pave the path for future studies on physics-inspired machine learning for scour prediction.
 
\section{Data Availability Statement}
\noindent
Some or all data used for this study are available from the corresponding author upon request. Models or codes generated or used during the study are proprietary or confidential in nature and may only be provided with restrictions. Related data and codes are available at \url{https://github.com/Data-Driven-Computational-Geotechnics/ScourSensePhase4/}.

\section{Acknowledgments}
\noindent
This research was supported by The University of Melbourne’s Spartan and the GEO\&CO Infrastructure HPC Center. The funding was provided by The University of Melbourne's Start-Up awarded to Dr Negin Yousefpour.

\bibliography{references/main}


\clearpage

\section*{Appendix}  \label{sec-appendix}
\addcontentsline{toc}{section}{Appendix}
\renewcommand{\thefigure}{A\arabic{figure}}
\renewcommand{\thetable}{A\arabic{table}}
\setcounter{figure}{0}
\setcounter{table}{0}
\noindent

\begin{longtable}{llllll}
    \label{tab:table_overall_performance} \\
    \caption{Comprehensive evaluation results of all experiments, comparing the performance of pure NN models and different SPINN variants across four bridge datasets.} \\
    \toprule
    Test Set & Base Model & Method & Physics-based Loss & Training Set (s) & Test MSE ($m^2$)  \\
    \midrule
    \endfirsthead
    \endhead
    212 & LSTM    & Pure NN & N/A   & 212 & 0.533±0.267 \\
    212 & LSTM    & Pure NN (General) & N/A   & All & 0.089±0.032 \\
    212 & LSTM    & SPINN   & HEC18 & 212 & 0.443±0.350 \\
    212 & LSTM    & SPINN   & TD    & 212 & 0.136±0.053 \\
    212 & LSTM    & SPINN   & GTD   & All & 0.144±0.049 \\
    212 & CNN     & Pure NN & N/A   & 212 & 0.142±0.022 \\
    212 & CNN     & Pure NN (General) & N/A   & All & 0.101±0.032 \\
    212 & CNN     & SPINN   & HEC18 & 212 & 0.150±0.039 \\
    212 & CNN     & SPINN   & TD    & 212 & 0.173±0.044 \\
    212 & CNN     & SPINN   & GTD   & All & 0.089±0.034 \\
    527 & LSTM    & Pure NN & N/A   & 527 & 0.141±0.074 \\
    527 & LSTM    & Pure NN (General) & N/A   & All & 0.113±0.037 \\
    527 & LSTM    & SPINN   & HEC18 & 527 & 0.135±0.091 \\
    527 & LSTM    & SPINN   & TD    & 527 & 0.084±0.035 \\
    527 & LSTM    & SPINN   & GTD   & All & 0.122±0.053 \\
    527 & CNN     & Pure NN & N/A   & 527 & 0.055±0.045 \\
    527 & CNN     & Pure NN (General) & N/A   & All & 0.123±0.021 \\
    527 & CNN     & SPINN   & HEC18 & 527 & 0.040±0.013 \\
    527 & CNN     & SPINN   & TD    & 527 & 0.050±0.024 \\
    527 & CNN     & SPINN   & GTD   & All & 0.130±0.037 \\
    539 & LSTM    & Pure NN & N/A   & 539 & 0.478±0.157 \\
    539 & LSTM    & Pure NN (General) & N/A   & All & 0.332±0.031 \\
    539 & LSTM    & SPINN   & HEC18 & 539 & 0.388±0.078 \\
    539 & LSTM    & SPINN   & TD    & 539 & 0.236±0.026 \\
    539 & LSTM    & SPINN   & GTD   & All & 0.325±0.037 \\
    539 & CNN     & Pure NN & N/A   & 539 & 0.283±0.069 \\
    539 & CNN     & Pure NN (General) & N/A   & All & 0.276±0.022 \\
    539 & CNN     & SPINN   & HEC18 & 539 & 0.285±0.070 \\
    539 & CNN     & SPINN   & TD    & 539 & 0.236±0.015 \\
    539 & CNN     & SPINN   & GTD   & All & 0.250±0.022 \\
    742 & LSTM    & Pure NN & N/A   & 742 & 0.884±0.589 \\
    742 & LSTM    & Pure NN (General) & N/A   & All & 0.325±0.021 \\
    742 & LSTM    & SPINN   & HEC18 & 742 & 0.388±0.036 \\
    742 & LSTM    & SPINN   & TD    & 742 & 0.260±0.021 \\
    742 & LSTM    & SPINN   & GTD   & All & 0.346±0.045 \\
    742 & CNN     & Pure NN & N/A   & 742 & 0.214±0.037 \\
    742 & CNN     & Pure NN (General) & N/A   & All & 0.310±0.040 \\
    742 & CNN     & SPINN   & HEC18 & 742 & 0.239±0.055 \\
    742 & CNN     & SPINN   & TD    & 742 & 0.233±0.033 \\
    742 & CNN     & SPINN   & GTD   & All & 0.335±0.024 \\
    \bottomrule
\end{longtable}

\begin{figure}[!h]
    \centering
    \includegraphics[width=1.0\linewidth]{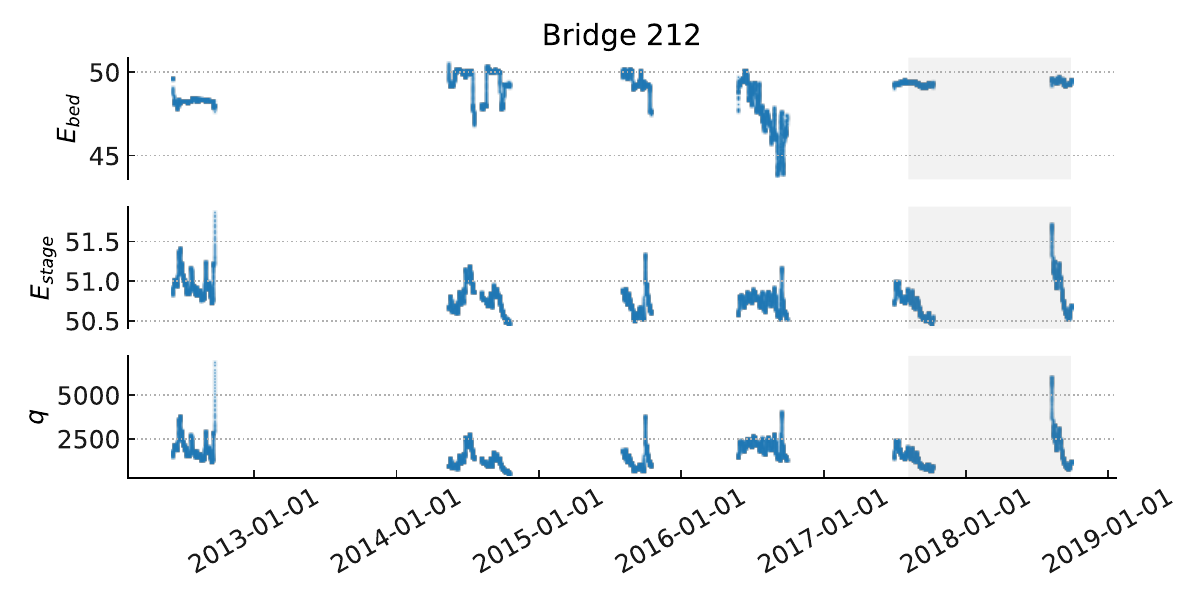}
    \caption{Historical time series data for Bridge 212 shows the three features: bed elevation ($m$), river stage elevation ($m$), and discharge ($m^3/s$). The grey zone shows the test subset, and the transparent part before is the training and validation subsets.}
    \label{fig:train_val_test_212}
\end{figure}

\begin{figure}[!h]
    \centering
    \includegraphics[width=1.0\linewidth]{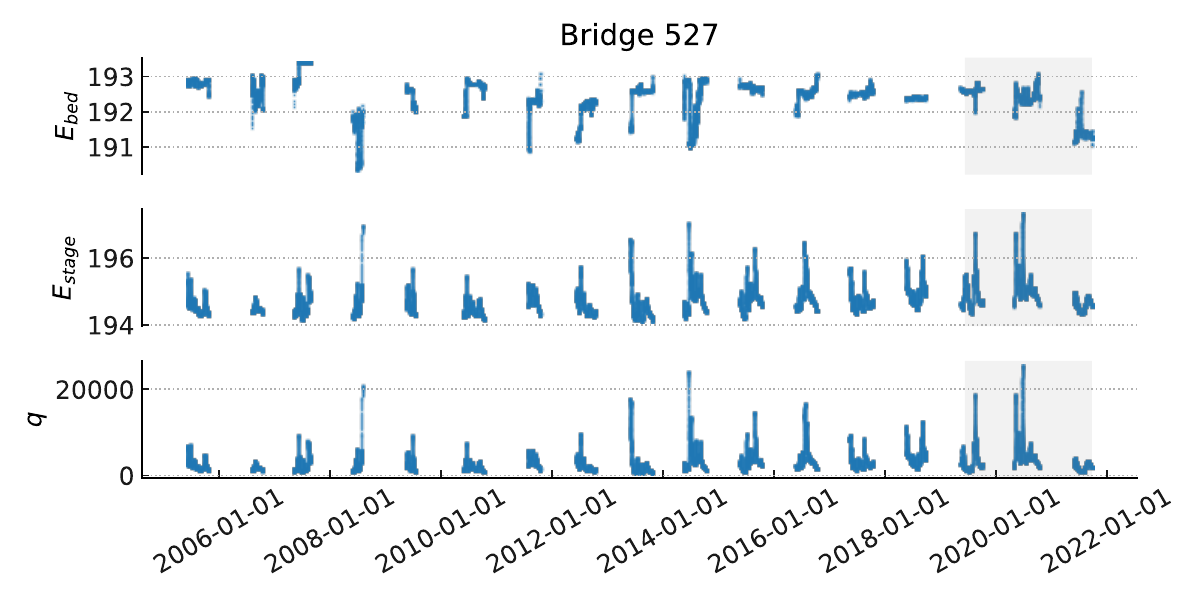}
    \caption{Historical time series data for Bridge 527 shows the three features: bed elevation ($m$), river stage elevation ($m$), and discharge ($m^3/s$). The grey zone shows the test subset, and the transparent part before is the training and validation subsets.}
    \label{fig:train_val_test_527}
\end{figure}

\begin{figure}[!ht]
    \centering
    \includegraphics[width=1.0\linewidth]{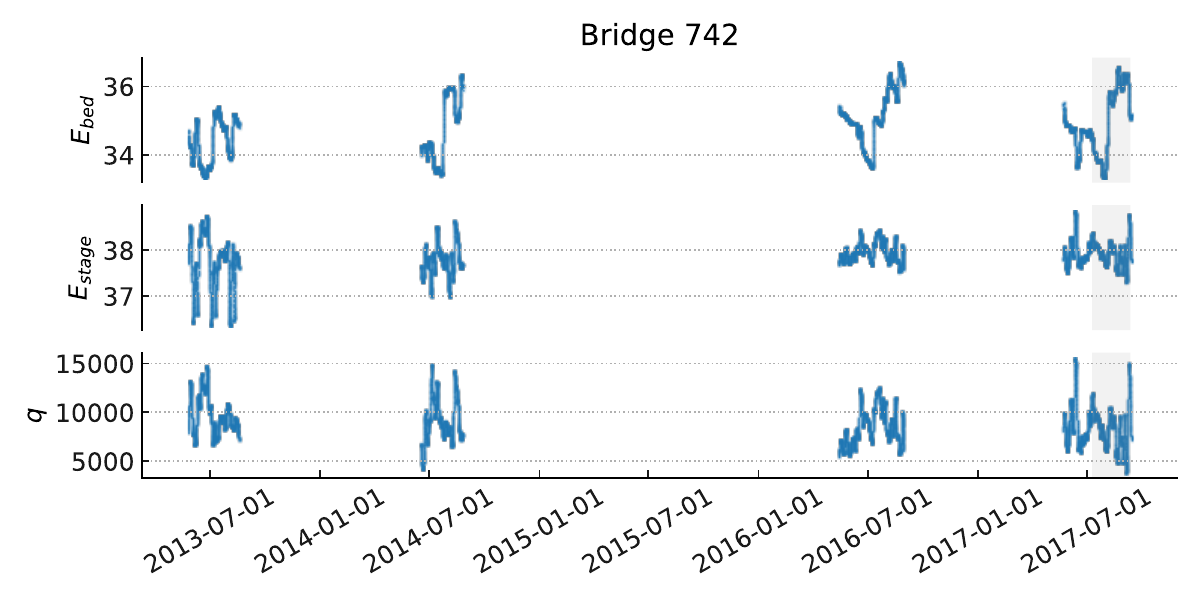}
    \caption{Historical time series data for Bridge 742 shows the three features: bed elevation ($m$), river stage elevation ($m$), and discharge ($m^3/s$). The grey zone shows the test subset, and the transparent part before is the training and validation subsets.}
    \label{fig:train_val_test_742}
\end{figure}

\begin{figure}
    \centering
    \includegraphics[width=0.9\linewidth]{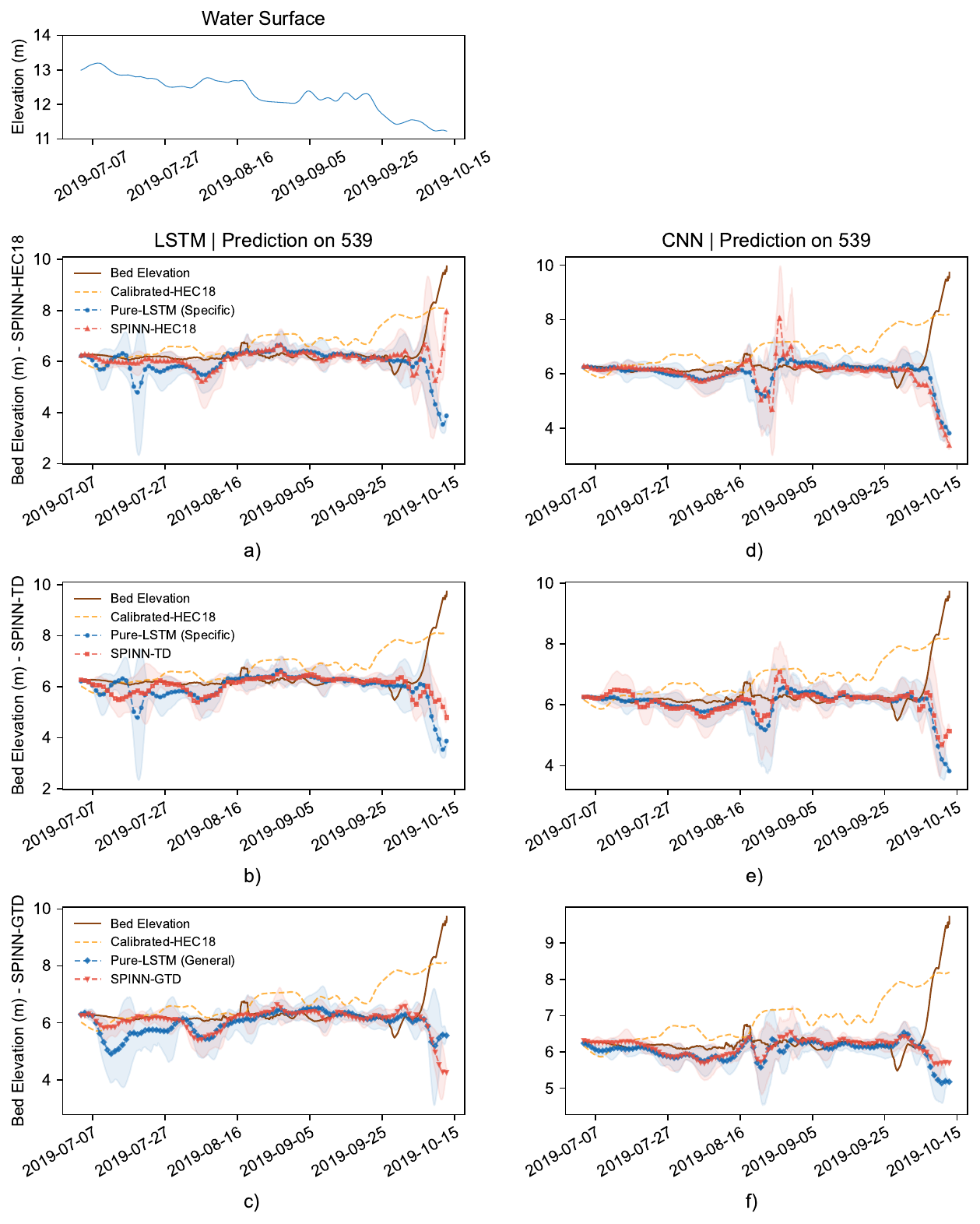}
    \caption{Model predictions versus actual observations for Bridge 539 test set (July-October 2019). Top graph displays the flow changes in this period. (a-c) show LSTM-based predictions, (d-f) display CNN-based predictions, and each row represents SPINN-HEC18, SPINN-TD, and SPINN-GTD, respectively.}
    \label{fig:539_test_2019}
\end{figure}

\begin{figure}
    \centering
    \includegraphics[width=0.9\linewidth]{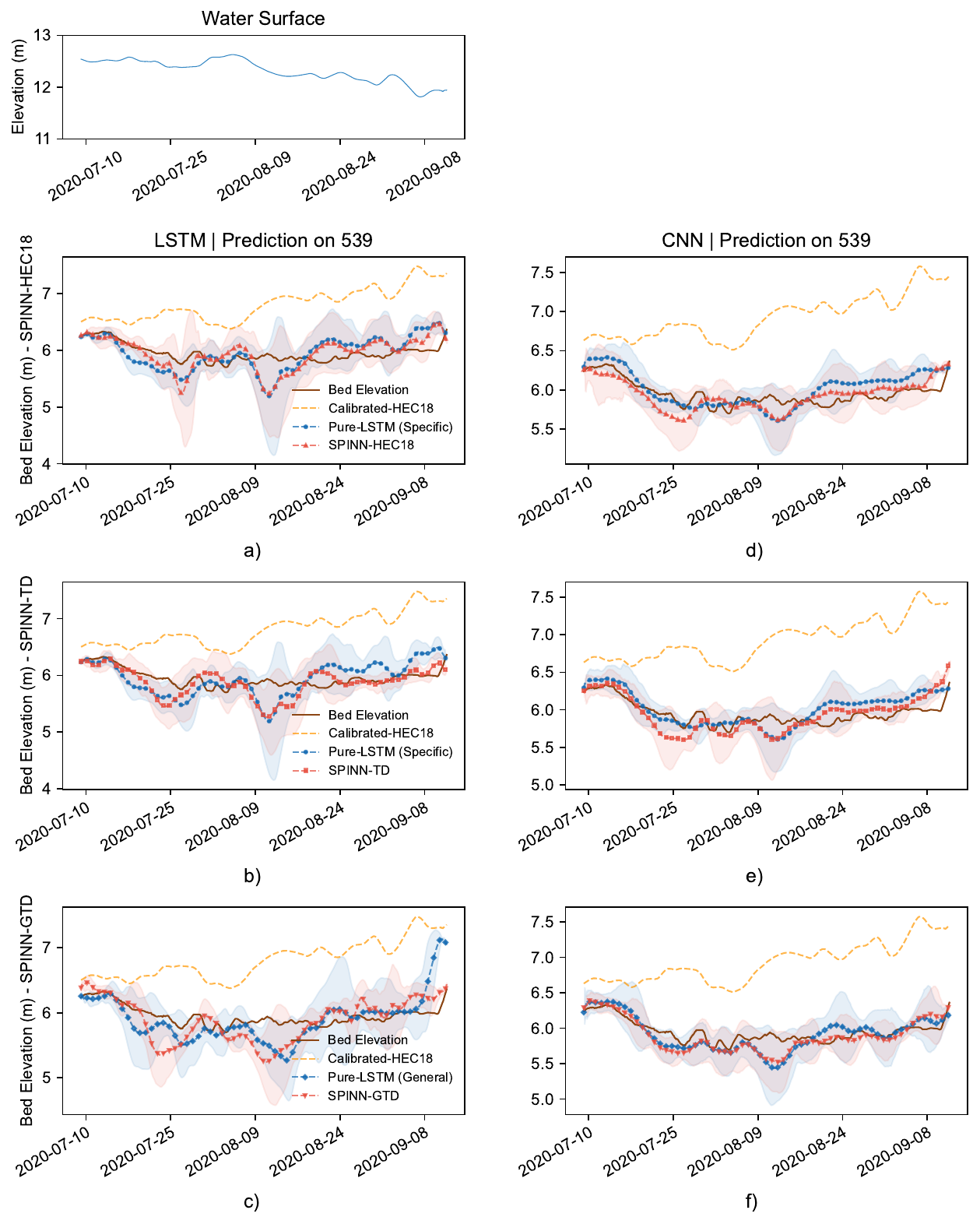}
    \caption{Model predictions versus actual observations for Bridge 539 test set (July-September 2020). Top graph displays the flow changes in this period. (a-c) show LSTM-based predictions, (d-f) display CNN-based predictions, and each row represents SPINN-HEC18, SPINN-TD, and SPINN-GTD, respectively.}
    \label{fig:539_test_2020}
\end{figure}

\end{document}